%% file: main.tex
\documentclass[runningheads]{llncs}
\usepackage{graphicx}
\usepackage{amsmath,amssymb} % define this before the line numbering.
\usepackage{color}
\usepackage[width=122mm,left=12mm,paperwidth=146mm,height=193mm,top=12mm,paperheight=217mm]{geometry}

\usepackage{times}
\usepackage{epsfig}
\usepackage{bm}
\usepackage{enumitem}
\usepackage{units}
\usepackage{fixltx2e}
\usepackage{afterpage}
\usepackage{microtype}
\usepackage{array}
\usepackage[flushleft]{threeparttable}
\usepackage{multirow}
\usepackage{booktabs}
\usepackage{booktabs,xcolor}
\usepackage{comment}
\usepackage{rotating}
\usepackage{subcaption}
\usepackage[colorlinks,bookmarks=false]{hyperref}

\definecolor{lightgray}{gray}{0.9}
\setlist{noitemsep} 

\newcommand{\tda}{\textsuperscript{\textdagger}} 
\newcommand{\tas}{\textsuperscript{\textasteriskcentered}} 
 
\newcommand{\tup}[1]{\textsuperscript{#1}}  

\newcommand{\tabg}[1]
{\includegraphics[width = 0.17 \textwidth]{img/f/#1}}
\newcommand{\tabr}[2]
{\multirow{#1}{*}{\rotatebox{90}{#2}}}
\newcommand{\tabrr}[1]
{\centering\rotatebox{90}{\centering\ #1}}

\begin{document}
% \linenumbers
\pagestyle{headings}
\mainmatter

\title{\textls[-3]{Sympathy for the Details: Dense Trajectories and Hybrid Classification Architectures for Action Recognition}}

\titlerunning{Sympathy for the Details: Dense Trajectories and Hybrid Classification ...
    %Architectures for Action Recognition
    }

\authorrunning{C\'{e}sar de Souza, Adrien Gaidon, Eleonora Vig, Antonio L\'{o}pez}

\author{C\'{e}sar Roberto de Souza\textsuperscript{1,2}, Adrien 
    Gaidon\textsuperscript{1}, Eleonora Vig\textsuperscript{3}, 
    Antonio Manuel L\'{o}pez\textsuperscript{2}}

\institute{
\textsuperscript{1}Computer Vision Group, Xerox Research Center Europe, Meylan, France \\
\textsuperscript{2}Centre de Visi\'{o} per Computador, Universitat Aut\`{o}noma de Barcelona, Bellaterra, Spain \\
\textsuperscript{3}German Aerospace Center, Wessling, Germany \\
{\tt\small \{cesar.desouza, adrien.gaidon\}@xrce.xerox.com, eleonora.vig@dlr.de, antonio@cvc.uab.es}
}

\maketitle
\begin{abstract}
Action recognition in videos is a challenging task due to the complexity of the
spatio-temporal patterns to model and the difficulty to acquire and learn on
large quantities of video data.
Deep learning, although a breakthrough for image classification and showing
promise for videos, has still not clearly superseded action recognition methods
using hand-crafted features, even when training on massive datasets.
In this paper, we introduce hybrid video classification architectures based on
carefully designed unsupervised representations of hand-crafted spatio-temporal
features classified by supervised deep networks.
As we show in our experiments on five popular benchmarks for action
recognition, our hybrid model combines the best of both worlds: it is data
efficient (trained on 150 to 10000 short clips) and yet improves significantly
on the state of the art, including recent deep models trained on millions of
manually labelled images and videos.
\end{abstract}

\input{abbreviations}

\let\thefootnote\relax\footnote{Accepted for publication in the 14th European Conference on Computer
    Vision (ECCV), Amsterdam, 2016.}

\input{introduction}
\input{related}
\input{shallow}

\input{hybrid}

\input{experiments}
\input{conclusion}

\bibliographystyle{splncs}
\bibliography{new}

\input{supp}

%% file: abbreviations.tex
\def\eg{\textit{e.g.,~}}
\def\cf{\textit{cf.~}}
\def\vs{\textit{vs.~}}
\def\ie{\textit{i.e.~}}
\def\wrt{\textit{w.r.t.~}}
\def\etal{\textit{et~al.~}}
\def\iid{\textit{i.i.d.~}}

\def\g{\gamma}
\def\d{\delta}
\def\p{\phi}
\def\s{\sigma}
\def\Re{\mathbb R}

% TODO AG: remove Baccouche2010 from table? (not state of the art + we don't do LSTM), maybe discuss below?
% TODO AG: any other method we could putin \tb ?
% TODO AG: add Evig2012?
\def\ta{\cite{Wang2011,Wang2013,Wang2013a,Wanga,Gaidon2014,Lan2014,Peng2014a,Hoai2014,Fernando2015}}  
\def\tb{\cite{Baccouche2010}, our method}  
\def\tc{\cite{Zha2015,Wang2015d,Tran2014,Wu2015a}}  
\def\td{\cite{Ji2013,Karpathy2014,Simonyan2014,Sun2015a,Wu2015,Donahue2015,Ng2015,Srivastava2015,Ballas2016,Gan2015,Wang2015,Feichtenhofer2016}}

%% file: introduction.tex
% !TEX root = main.tex

\section{Introduction}\label{sec:intro}

Classifying human actions in real-world videos is an open research problem with
many applications in multimedia, surveillance, and robotics~\cite{Vrigkas2015}.
Its complexity arises from the variability of imaging conditions, motion,
appearance, context, and interactions with persons, objects, or
the environment over different spatio-temporal extents.
Current state-of-the-art algorithms for action recognition are based on
statistical models learned from manually labeled videos. They belong to two
main categories: models relying on features \emph{hand-crafted} for action
recognition (\eg~\ta), or more recent end-to-end \emph{deep architectures}
(\eg~\td).
These approaches have complementary strengths and weaknesses.
Models based on hand-crafted features are data efficient, as they can easily
incorporate structured prior knowledge (\eg the importance of motion boundaries
along dense trajectories~\cite{Wang2011}), but their lack of flexibility may
impede their robustness or modeling capacity.
Deep models make fewer assumptions and are learned end-to-end from data
(\eg~using 3D-ConvNets~\cite{Tran2014}), but they rely on hand-crafted
architectures and the acquisition of large manually labeled video datasets
(\eg Sports-1M~\cite{Karpathy2014}), a costly and error-prone
process that poses optimization, engineering, and infrastructure challenges.

Although deep learning for videos has recently made significant improvements
(\eg~\cite{Simonyan2014,Tran2014,Sun2015a}), models using hand-crafted features
are the state of the art on many standard action recognition benchmarks
(\eg~\cite{Lan2014,Fernando2015,Hoai2014}).
These models are generally based on {\em improved Dense Trajectories} (iDT)
\cite{Wang2013,Wang2013a} with Fisher Vector (FV)
encoding~\cite{Perronnin2007,Perronnin2010}.
% of local spatio-temporal descriptors (trajectory coordinates, HOG, HOF, MBH)
% computed from RGB and optical flow inputs.
%
Recent deep models for action recognition therefore combine their predictions
with complementary ones from iDT-FV for better
performance~\cite{Tran2014,Wang2015d}.

\begin{figure}[t]
\begin{center}
\includegraphics[page=1,width=12cm]{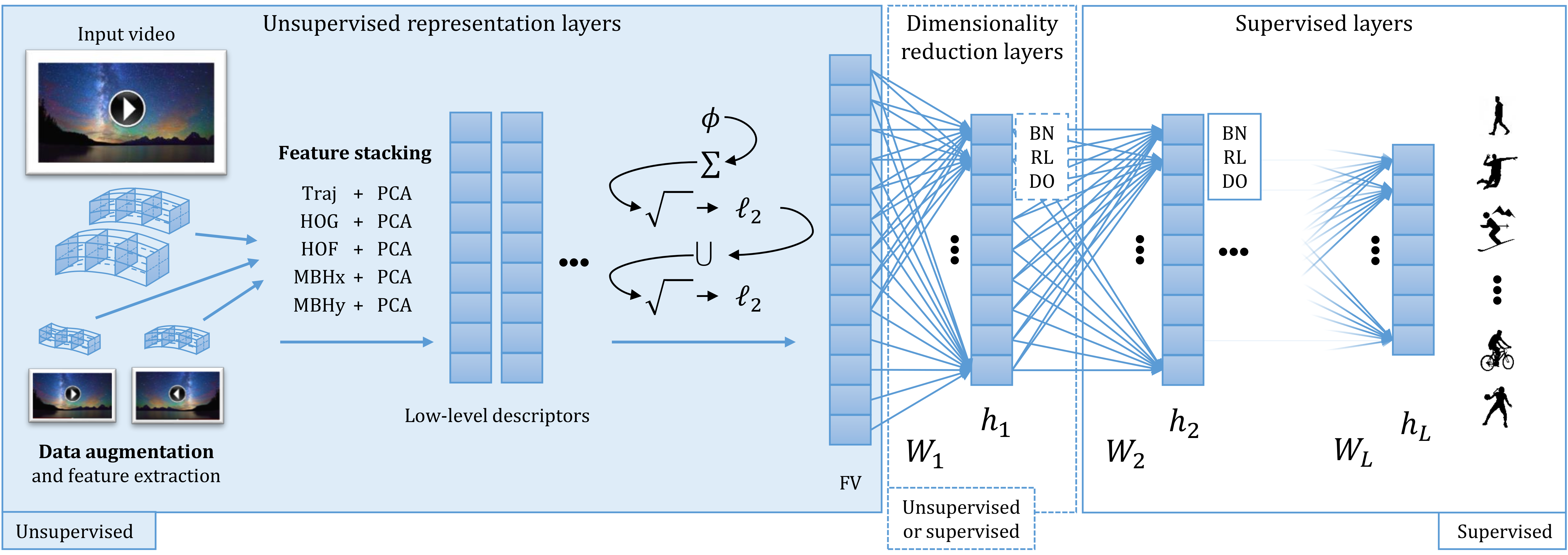}
\vspace{-0.4cm}
\parbox{12cm}{
\vspace{0.1cm}
\scriptsize
Hand-crafted features are extracted along optical flow trajectories from 
original and generated videos. Those features are then normalized using 
RootSIFT~\cite{Arandjelovi2012}, PCA-transformed, and augmented with 
their $(x,y,t)$ coordinates, forming our low-level descriptors. The 
descriptors for each feature channel are then encoded ($\phi$) as Fisher 
Vectors, separately aggregated ($\Sigma$) into a video-level 
representation, square-rooted, and $\ell_2$-normalized. These 
representations are then concatenated ($\cup$) and renormalized. A 
dimensionality reduction layer is learned supervisedly or 
unsupervisedly. Supervised layers are followed by Batch-Normalization 
(BN)~\cite{Ioffe2015}, ReLU (RL) non-linearities~\cite{Nair2010}, and 
Dropout (DO)~\cite{Hinton2014} during training. The last layer uses
sigmoids (multi-label datasets) or softmax (multi-class datasets)
non-linearities to produce action-label estimates.}

\end{center}
\vspace{-0.1cm}
\caption{Our hybrid unsupervised and supervised deep multi-layer architecture.}
\label{fig:arch}
\vspace{-0.6cm}
\end{figure}

In this paper, we study an alternative strategy to \emph{combine the best of
both worlds via a single hybrid classification architecture} consisting in
chaining sequentially the iDT hand-crafted features, the unsupervised FV
representation, unsupervised or supervised dimensionality reduction, and
a supervised deep network (\cf~Figure~\ref{fig:arch}).
This family of models was shown by Perronnin and Larlus \cite{Perronnin2015} to
perform on par with the deep convolutional network of
Krizhevsky~\etal~\cite{Krizhevsky2012} for large scale image classification.
We adapt this type of architecture differently for action recognition in videos
with particular care for data efficiency.

Our {\bf first contribution} consists in a careful design of the first
\emph{unsupervised} part of our hybrid architecture, which even with a simple
SVM classifier is already on par with the state of the art.
We experimentally observe that showing \emph{sympathy for the details}
(\eg~spatio-temporal structure, normalization) and doing \emph{data
augmentation by feature stacking} (instead of duplicating training samples) are
critical for performance, and that optimal design decisions generalize across
datasets.

Our {\bf second contribution} consists in a \emph{data efficient hybrid
architecture} combining unsupervised representation layers with a
deep network of multiple fully connected layers. We show that \emph{supervised
mid-to-end learning of a dimensionality reduction layer together with
non-linear classification layers} yields an excellent compromise between
recognition accuracy, model complexity, and transferability of the model across
datasets thanks to reduced risks of overfitting and modern optimization
techniques.

The paper is organized as follows. Section \ref{sec:related} reviews the
related works in action recognition. Section \ref{sec:shallow} presents the
details of the first unsupervised part (based on iDT-FV) of our hybrid model,
while Section \ref{sec:hybrid} does so for the rest of the architecture and our
learning algorithm.  In Section \ref{sec:experiments} we report experimental
conclusions from parametric studies and comparisons to the state of the art on
five widely used action recognition datasets of different sizes. 
%
% TODO: Remove if need space?
In particular, we show that our hybrid architecture improves significantly upon
the current state of the art, including recent combinations of iDT-FV predictions
with deep models trained on millions of images and videos.

%% file: related.tex
% !TEX root = main.tex

\section{Related Work}\label{sec:related}

%\vspace{-2mm}
Existing action recognition approaches (\cf~\cite{Vrigkas2015} for a recent
survey) can be organized into four broad categories based on whether they
involve \textit{hand-crafted} \vs \textit{deep-based} video features, and a
\textit{shallow} \vs \textit{deep} classifier, as summarized in
Table~\ref{tab:related}.

% XXX ta, tb, tc, and td are defined in abbreviations.tex

\makeatletter
\newcommand{\thickhline}{%
    \noalign {\ifnum 0=`}\fi \hrule height 1.2pt
    \futurelet \reserved@a \@xhline
}
\newcolumntype{"}{@{\hskip\tabcolsep\vrule width 1.2pt\hskip\tabcolsep}}
\makeatother

% Font sizes of headings. Table captions should always be positioned above the
% tables. The nal sentence of a table caption should end without a full stop

\begin{table}[h]
 \vspace{-9mm}
 \caption{Categorization of related recent action recognition methods}
  \centering
 \scriptsize
 \setlength\tabcolsep{5pt} % default value: 6pt
 \begin{tabular}{|l"c|c|}
     \hline
                                & \sc{Shallow classifier} & \sc{Deep classifier}\\\thickhline
     \sc{Hand-crafted features} & \ta & \tb \\\hline
     \sc{Deep-based features}   & \tc & \td \\\hline
  \end{tabular}
\label{tab:related}
\vspace{-5mm}
\end{table}

\noindent \textbf{Hand-crafted features, shallow classifier.}
A significant part of the progress on action recognition is driven by the
development of local hand-crafted spatio-temporal features encoded as
bag-of-words representations classified by ``shallow'' classifiers
such as SVMs~\ta .
%~\cite{Wang2011,Lan2014,Peng2014b,Wang2013,Wang2013a,Peng2014a,Wanga,Fernando2015}.
%
Most successful approaches use \emph{improved Dense Trajectories
(iDT)}~\cite{Wang2013} to aggregate local appearance and motion descriptors
into a video-level representation through the Fisher Vector (FV)
encoding~\cite{Perronnin2007,Perronnin2010}.
Local descriptors such as HOG~\cite{N.Dalal2005}, HOF~\cite{Dalal2006}, and 
MBH~\cite{Wang2011} are extracted along dense point trajectories obtained from
optical flow fields.
There are several recent improvements to iDT, for instance, using motion
compensation~\cite{Wanga,Gaidon2012,Jain2013,Gaidon2014} and stacking of FVs to
obtain a multi-layer encoding similar to mid-level
representations~\cite{Peng2014b}.
To include global spatio-temporal location information, Wang \etal~\cite{Wanga}
compute FVs on a spatio-temporal pyramid (STP)~\cite{Laptev2008} and use
Spatial Fisher Vectors (SFV)~\cite{Krapac2011}.
Fernando~\etal~\cite{Fernando2015} model the global temporal evolution over the
entire video using ranking machines learned on time-varying average FVs.
Another recent improvement is the Multi-skIp Feature Stacking (MIFS)
technique~\cite{Lan2014}, which stacks features extracted at multiple
frame-skips for better invariance to speed variations.
An extensive study of the different steps of this general iDT pipeline and
various feature fusion methods is provided in~\cite{Peng2014a}.
% VideoDarwin?

\noindent \textbf{End-to-end learning: deep-based features, deep classifier.}
The seminal supervised deep learning approach of
Krizhevsky~\etal~\cite{Krizhevsky2012} has enabled impressive performance
improvements on large scale image classification benchmarks, such as
ImageNet~\cite{Russakovsky2014}, using Convolutional Neural Networks
(CNN)~\cite{Lecun1989}.
Consequently, several approaches explored deep architectures for action
recognition.
While earlier works resorted to unsupervised learning of 3D spatio-temporal
features~\cite{Le2011}, supervised end-to-end learning has recently gained
popularity~\td.
%The typical input to these network is a stack of consecutive frames and 
Karpathy~\etal~\cite{Karpathy2014} studied several architectures and fusion
schemes to extend 2D CNNs to the time domain. Although trained on the very
large Sports-1M dataset, their 3D networks performed only marginally better
than single-frame models.
% and are outperformed by iDT-type methods~\cite{ }. % removed: true for all deep
%
To overcome the difficulty of learning spatio-temporal features jointly, the
Two-Stream architecture~\cite{Simonyan2014} is composed of two CNNs trained
independently, one for appearance modeling on RGB input, and another for
temporal modeling on stacked optical flow. Sun \etal~\cite{Sun2015a} factorize
3D CNNs into learning 2D spatial kernels, followed by 1D temporal ones. 
Alternatively, other recent works use recurrent neural networks (RNN) in
conjunction with CNNs to encode the temporal evolution of
actions~\cite{Donahue2015,Ng2015,Ballas2016}.
%
% LRCN~\cite{Donahue2015} processes video frames through the Two-Stream CNN
% (2S-CNN) whose outputs are fed into a stack of LSTMs.  \cite{Ng2015} proposes
% different temporal feature-pooling architectures to combine information over
% longer time periods. Most recently, Ballas \etal~\cite{Ballas2016} makes use of
% convolutional features from different layers of a pre-trained net as input to a
% GRU-RNN.
%
% In a different line of work, \cite{Srivastava2015} investigated the
% unsupervised learning of video representations using LSTMs.
%
Overall, due to the difficulty of training 3D-CNNs and the need for vast
amounts of training videos (\eg Sports-1M~\cite{Karpathy2014}), end-to-end
methods report only marginal improvements over traditional baselines, and
our experiments show that the iDT-FV often outperforms these
approaches.

\noindent \textbf{Deep-based features, shallow classifier.}
Several works~\tc~explore the encoding of general-purpose deep-learned features
in combination with ``shallow'' classifiers, transferring ideas from the iDT-FV
algorithm.
Zha~\etal~\cite{Zha2015} combine CNN features trained on ImageNet~\cite{Russakovsky2014}
with iDT features through a Kernel SVM.
The TDD approach~\cite{Wang2015d} extracts per-frame convolutional feature maps
from two-stream CNN~\cite{Simonyan2014} and pools these over spatio-temporal
cubes along extracted trajectories. Similar to~\cite{Karpathy2014},
C3D~\cite{Tran2014} learns general-purpose features using a 3D-CNN, but the
final action classifier is a linear SVM.
% EV removed below b/c of space constraints:
%RNNs have also been explored in this category. For video captioning, Yao
%\etal~\cite{Yao2015} proposed a 3D CNN-RNN encoder-decoder architecture. 
%
%This encodes the input using a 3D-CNN trained on space-time cuboids represented by histograms (HOG, HOF, MBH), and decodes this with an LSTM and frame-level soft-attention. 
%Wu \etal's~\cite{Wu2015a} hybrid architecture feeds Two-Stream features into
%two separate LSTMs and an additional fusion network performs the video-level
%feature fusion and classification.
Like end-to-end deep models, these methods rely on large datasets to learn
generic useful features, which in practice perform on par or worse than iDT.
% XXX AG last statement above a bit violent, but we need conclu/transi

\noindent \textbf{Hybrid architectures: hand-crafted features, deep classifier.}
There is little work on using unsupervised encodings of hand-crafted
local features in combination with a deep classifier.
In early work, Baccouche \etal~\cite{Baccouche2010} learn temporal dynamics of
traditional per-frame SIFT-BOW features using a RNN.  The method, coupled with
camera motion features, improves on BoW-SVM for a small set of soccer videos.
% XXX AG: add more of this type? maybe Yao2015?

Our work lies in this category, as it combines the strengths of iDT-FV encodings
and supervised deep multi-layer non-linear classifiers.
Our method is inspired by the recently proposed hybrid image classification
architecture of Perronnin and Larlus~\cite{Perronnin2015}, who stack several
unsupervised FV-based and supervised layers. Their hybrid architecture shows
significant improvements over the standard FV pipeline, closing the gap
on~\cite{Krizhevsky2012}, which suggests there is still much to learn about
FV-based methods.

Our work investigates this type of hybrid architectures, with several
noticeable differences: (i) FV is on par with the current state of the art for
action recognition, (ii) iDT features contain many different appearance and
motion descriptors, which also results in more diverse and higher-dimensional
FV, (iii) most action recognition training sets are small due to the cost of
labeling and processing videos, so overfitting and data efficiency are major
concerns.
In this context, we adopt different techniques from modern hand-crafted and
deep models, and perform a wide architecture and parameter study showing
conclusions regarding many design choices specific to action recognition. 

%% file: shallow.tex
% !TEX root = main.tex

\section{Fisher Vectors in Action: From Baseline to State of the Art}\label{sec:shallow}

We first recall the iDT approach of Wang \& Schmid~\cite{Wang2013}, then
describe the improvements that can be stacked together to transform this strong
baseline into a state-of-the-art method for action recognition. In particular,
we propose a data augmentation by feature stacking method motivated by
MIFS~\cite{Lan2014} and data augmentation for deep models.

\subsection{Improved Dense Trajectories}\label{ss:idt}
% goal : describe everything related to iDT as reproduced in first row of Table 2

\noindent \textbf{Local spatio-temporal features.}
% all the details about flow + Traj + descs + PCA + motion compensation
The iDT approach used in many state-of-the-art action
recognition algorithms
(\eg~\cite{Wang2013,Wang2013a,Wanga,Lan2014,Peng2014b,Peng2014a,Fernando2015}) 
consists in first extracting dense trajectory video features~\cite{Wang2011}
that efficiently capture appearance, motion, and spatio-temporal statistics. 
Trajectory shape (Traj)~\cite{Wang2011}, HOG \cite{N.Dalal2005}, HOF
\cite{Dalal2006}, and MBH \cite{Wang2011} descriptors are extracted along
trajectories obtained by median filtering dense optical flow. We extract dense
trajectories from videos in the same way as in \cite{Wang2013}, 
% Note: should we cite HOG, HOF, MBH again? Its already done in related works
%
applying RootSIFT normalization~\cite{Arandjelovi2012} ($\ell_1$
normalization followed by square-rooting) to all descriptors.
The number of dimensions in the Traj, HOG, HOF, MBHx, and MBHy descriptors
are respectively 30, 96, 108, 96, and 96.

\noindent \textbf{Unsupervised representation learning.}
Before classification, we combine the multiple trajectory descriptors in a
single video-level representation by accumulating their Fisher Vector encodings
(FV)~\cite{Perronnin2007,Perronnin2010}, which was shown to be particularly
effective for action recognition~\cite{Wanga,Chatfield}.
This high-dimensional representation is based on the gradient of a generative
model, a Gaussian Mixture Model (GMM), learned in an \emph{unsupervised} manner
on a large set of trajectory descriptors in our case.
%
% XXX below from Florent & Diane's, remove if need space, but try to keep
% (Florent has it + pimps the paper a bit)
Given a GMM with $K$ Gaussians, each parameterized by its mixture weight $w_k$,
mean vector $\mu_k$, and standard deviation vector $\s_k$ (assuming a diagonal
covariance), the FV encoding of a trajectory descriptor $x \in \Re^D$ is
$\Phi(x)=[\p_1(x), \ldots, \p_K(x)] \in \Re^{2KD}$, where:

\begin{equation}
  \p_k(x)=\left[ \frac{\g(k)}{\sqrt{w_k}} \left(\frac{x-\mu_k}{\s_k} \right), \frac{\g(k)}{\sqrt{2w_k}} \left(\frac{(x-\mu_k)^2}{\s_k^2}-1\right)\right]
\end{equation}

\noindent and $\g(k)$ denotes the soft assignment of descriptor $x$ to Gaussian $k$.
We use $K=256$ Gaussians as a good compromise between accuracy and efficiency
\cite{Wang2013,Wang2013a,Wanga}.
%
% XXX ~\cite{Wang2011,Jain2013}
We randomly sample 256,000 trajectories from the pool of training videos,
irrespectively of their labels, to learn one GMM per descriptor channel using 10
iterations of EM.
Before learning the GMMs, we apply PCA to the descriptors, reducing their
dimensionality by a factor of two.
The reduced dimensions for the are, respectively, 15, 48, 54, 48, 48.
After learning the GMMs, we extract FV encodings for all descriptors in each
descriptor channel and combine these encodings into a per-channel, video-level
representation using sum-pooling, \ie by adding FVs together before normalization. 
In addition, we apply further post-processing and normalization steps, as
discussed in the next subsection.

\noindent \textbf{Supervised classification.}
When using a linear classification model, we use a linear SVM. As it is standard 
practice and in order to ensure comparability with previous works \cite{Wang2013,Lan2014,Wang,Wang2015d}, 
we fix $C = 100$ unless stated otherwise and use \textit{one-vs-rest} for multi-class and multi-label 
classification. This forms a strong baseline for action recognition, as shown by 
previous works \cite{Wanga,Wang2015d} and confirmed in our experiments. 
We will now show how to make this baseline competitive with recent
state-of-the-art methods.

\subsection{Bag of Tricks for Bag-of-Words}\label{ss:idtricks}
% one sub-section per trick in rows 2,3,4 + last one about our trick

%One of the reasons for the resilience of this method is the steady increase in
%the number of improvements proposed since its
%invention~\cite{Perronnin2010,Chatfield2014,Chatfield,Lan2014,Peng2014a,Wang2015d}.

\noindent \textbf{Incorporating global spatio-temporal structure.}
Incorporating the spatio-temporal position of local features can improve the FV
representation.
We do not use spatio-temporal pyramids (STP)~\cite{Laptev2008}, as they
significantly increase both the dimensionality of the representation and its
variance~\cite{Jorge}.
Instead, we simply concatenate the PCA-transformed descriptors with their
respective $(x,y,t) \in \mathbb{R}^3$ coordinates, as in \cite{Jorge,Lan2014}.
We refer to this method as Spatio-Temporal Augmentation (STA).
This approach is linked to the Spatial Fisher Vector (SFV)~\cite{Krapac2011}, a
compact model related to soft-assign pyramids, in which the descriptor
generative model is extended to explicitly accommodate the $(x,y,t)$
coordinates of the local descriptors.
When the SFV is created using Gaussian spatial models (\cf~eq. 18 in
\cite{Krapac2011}), the model becomes equivalent to a GMM created from
augmented descriptors (assuming diagonal covariance matrices).
Using STA, the dimensions for the descriptors before GMM estimation become 
18, 51, 57, 51, 51. With 256 Gaussians, the dimension of the FVs generated 
for each descriptor channel are therefore 9,216, 26,112, 29,184, 26,112, 
and 26,112.

\noindent \textbf{Normalization.}
% Another important detail to obtain maximum performance with bag-of-words models
% is their normalization.
%
We apply signed-square-rooting followed by $\ell_2$ normalization, then
concatenate all descriptor-specific FVs and reapply this same normalization,
following \cite{Lan2014}. The double normalization re-applies square rooting,
and is thus equivalent to using a smaller power normalization
\cite{Perronnin2010}, which improves action recognition
performance~\cite{Narayan2015}.

\noindent \textbf{Multi-Skip Feature Stacking (MIFS).}
MIFS~\cite{Lan2014} improves the robustness of FV to videos of different
lengths by increasing the pool of features with frame-skipped versions of the
same video. Standard iDT features are extracted from those frame-skipped
versions and stacked together before descriptor encoding, decreasing the 
expectation and variance of the condition number \cite{Lan2014,Poggio2003,Bousquet2002}
of the extracted feature matrices.
We will now see that the mechanics of this technique can be expanded to other
transformations.

\subsection{Data Augmentation by Feature Stacking (DAFS)}\label{ss:feataug}

Data augmentation is an important part of deep learning~
\cite{Wang2015d,Chatfield2014,Vincent2016}, but it is rarely used with
hand-crafted features and shallow classifiers, particularly for action
recognition where duplicating training examples can vastly increase the
computational cost.
Common data augmentation techniques for images include the use of random
horizontal flipping \cite{Chatfield2014,Wang2015d}, random cropping
\cite{Chatfield2014}, and even automatically determined transformations
\cite{Paulin2014}.
For video classification, \cite{Hoai2014,Fernando2015} duplicate the training
set by mirroring.

Instead, we propose to generalize MIFS to arbitrary transformations, an
approach we call \emph{Data Augmentation by Feature Stacking} (DAFS).
First, we extract features from multiple transformations of an input video
(frame-skipping, mirroring, etc.) that do not change its semantic category.
Second, we obtain a large feature matrix by stacking the obtained
spatio-temporal features prior to encoding.
Third, we encode the feature matrix, pool the resulted encodings, and apply the
aforementioned normalization steps along this pipeline to obtain a \emph{single
augmented video-level representation}.

This approach yields a representation that simplifies the learning problem, as it
can improve the condition number of the feature matrix further than just MIFS
thanks to leveraging data augmentation techniques traditionally used for deep
learning.
In contrast to data augmentation for deep approaches, however, we build a single
more robust and useful representation instead of duplicating training examples.
Note also that DAFS is particularly suited to FV-based representation of videos
as pooling FV from a much larger set of features decreases one of the sources
of variance for FV~\cite{Boureau2010}.

After concatenation, the final representation for each video is 116,736-dimensional.

%% file: hybrid.tex
% !TEX root = main.tex

\section{Hybrid Classification Architecture for Action Recognition}\label{sec:hybrid}
% goal: describe methods from table 3, Figure 2 + SDR

\subsection{System Architecture}

Our hybrid action recognition model combining FV with neural networks (cf.
Fig.~\ref{fig:arch}) starts with the previously described steps of our
iDT-DAFS-FV pipeline, which can be seen as a set of \emph{unsupervised layers}.
\newcommand{\matr}[1]{#1}       % for marking matrices (disabled)   
The next part of our architecture consists of a set of $L$ fully connected
\emph{supervised layers}, each comprising a dot-product followed by a
non-linearity.
Let $\matr{h_{0}}$ denote the FV output from the last unsupervised layer in our
hybrid architecture, $\matr{h_{j-1}}$ the input of layer $j \in \{1, ..., L\}$,
$\matr{h_{j}} = g(\matr{W_j} \matr{h_{j-1}})$ its output, with $\matr{W_j}$ the
corresponding parameter matrix to be learned.
We omit the biases from our equations for better clarity.
For intermediate hidden layers we use the Rectified Linear Unit (ReLU)
non-linearity~\cite{Nair2010} for $g$.
For the final output layer we use different non-linearity functions depending
on the task. For multi-class classification over $c$ classes, we use the
softmax function $g(\matr{z_i}) = \exp(z_i)/\sum_{k=1}^c exp(z_k)$. For
multi-label tasks we consider the sigmoid function
$g(z_i) = 1/(1 + exp(-z_i))$. 

Connecting the last unsupervised layer to the first supervised layer
can result in a much higher number of weights in this section than in all other
layers of the architecture. Since this might be an issue for small datasets due
to the higher risk of overfitting, we study the impact of different ways to
learn the weights of this \emph{dimensionality reduction layer}: either with
unsupervised learning (\eg using PCA as in~\cite{Perronnin2015}), or by learning
a low-dimensional projection end-to-end with the next layers of the
architecture.

\subsection{Learning}

\noindent \textbf{Unsupervised layers.}
Our unsupervised layers are learned as described in Section~\ref{ss:idt}.
Namely, we learn one GMM of $K=256$ Gaussians per descriptor channel using EM
on a set of 256,000  trajectories randomly sampled from the pool of training
videos. 

\noindent \textbf{Supervised layers.}
We use the standard cross-entropy between the network output $\hat{y}$ and the
ground-truth label vectors $y$ as loss function. For multi-class classification
problems, we minimize the categorical cross-entropy cost function over all $n$
samples:

\begin{equation}
C_{cat}(y, \hat{y}) = - \sum_{i=1}^n \sum_{k=1}^{c} y_{ik} log(\hat{y}_{ik}),
\end{equation}

\noindent whereas for multi-label problems we minimize the binary cross-entropy:

\begin{equation}
C_{bin}(y, \hat{y}) = - \sum_{i=1}^n \sum_{k=1}^{c} y_{ik} log(\hat{y}_{ik}) - (1-y_{ik}) log(1-\hat{y}_{ik}).
\end{equation}

\noindent \textbf{Optimization.}
For parameter optimization we use the recently introduced Adam algorithm
\cite{Kingma2014}. Since Adam automatically computes individual adaptive
learning rates for the different parameters of our model, this alleviates the
need for fine-tuning of the learning rate with a costly grid-search or
similar methods.

Adam uses estimates of the first and second-order moments of the gradients in
the update rule:
\begin{equation}
\theta_t \leftarrow \theta_{t-1} - \alpha \cdot \frac{m_t}{(1-\beta_1^t)\sqrt{\frac{v_t}{1-\beta_2^t}}+\epsilon}
\text{\quad where \quad}
\begin{matrix}
    g_t & \leftarrow & \nabla_{\theta}f(\theta_{t-1}) \\
    m_t & \leftarrow & \beta_1 \cdot m_{t-1} + (1 - \beta_1) \cdot g_t \\
    v_t & \leftarrow & \beta_2 \cdot v_{t-1} + (1 - \beta_2) \cdot {g_t}^2 \\
\end{matrix}
\end{equation}

\noindent and where $f(\theta)$ is the function with parameters $\theta$ to be 
optimized, $t$ is the index of the current iteration, $m_0 = 0$, $v_0 = 0$, 
and $\beta_1^t$ and $\beta_2^t$ denotes $\beta_1$ and $\beta_2$ to 
the power of $t$, respectively. We use the default values for its parameters $\alpha = 0.001$,
$\beta_1 = 0.9$, $\beta_2~=~0.999$, and $\epsilon = 10^{-8}$ 
proposed in \cite{Kingma2014} and implemented in Keras~\cite{Chollet2015}.

\noindent \textbf{Batch normalization and regularization.}
During learning, we use batch normalization (BN)~\cite{Ioffe2015} and dropout
(DO) \cite{Hinton2014}.
Each BN layer is placed immediately before the ReLU non-linearity and
parametrized by two vectors $\gamma$ and $\beta$ learned alongside each
fully-connected layer. Given a training set $X = \{x_1, x_2, ..., x_n\}$ of $n$ 
training samples, the transformation learned by BN for each input vector $x \in X$
is given by:

\begin{equation}
	BN(x; \gamma, \beta) = \gamma \frac{x - \mu_{B}}{\sqrt{\sigma_{B}^2 + \epsilon}} + \beta
	\text{\quad where \quad}
    \begin{matrix}
    	\mu_{B}      \leftarrow \frac{1}{n} \displaystyle\sum_{i=1}^n x_i \\ 
    	\sigma_{B}^2 \leftarrow \frac{1}{n} \displaystyle\sum_{i=1}^n (x_i - \mu_{B})^2 \\
    \end{matrix}
\end{equation}

\noindent Together with DO, the operation performed by hidden layer $j$ can now be
expressed as $h_{j} = r \odot g(BN(W_j h_{j-1}; \gamma_j, \beta_j))$, where $r$ is a
vector of Bernoulli-distributed variables with probability $p$ and $\odot$ denotes
the element-wise product.
We use the same DO rate $p$ for all layers.
The last output layer is not affected by this modification. 

%Finally, we always consider the same dropout rate for all layers; the 
%same number of neurons in each supervised layer; and we always use 
%Glorot \cite{Glorot2010} uniform initialization when initializing the 
%network's weights. 

\noindent \textbf{Dimensionality reduction layer.}
When unsupervised, we fix the weights of the dimensionality reduction layer
from the projection matrices learned by PCA dimensionality reduction
followed by whitening and $\ell_2$ normalization~\cite{Perronnin2015}.  When it
is supervised, it is treated as the first fully-connected layer, to which we
apply BN and DO as with the rest of the supervised layers. 
To explain our initialization strategy for the unsupervised case, let us 
denote the set of $n$ mean-centered $d$-dimensional FVs for each sample 
in our dataset as a matrix $X \in \mathbb{R}^{d \times n}$. Recall that 
the goal of PCA projection is to find a $r \times d$ transformation 
matrix $P$, where $r \leqslant d$ on the form $Z = PX$ such that the 
rows of $Z$ are uncorrelated, and therefore its $d \times d$ scatter 
matrix $S = Z{Z}^t$ is diagonal. In its primal form, this can be 
accomplished by the diagonalization of the $d \times d$ covariance 
matrix $X X^t$. However, when $n \ll d$ it can become computationally 
inefficient to compute $X X^t$ explicitly. For this reason, we 
diagonalize the $n \times n$ Gram matrix $X^tX$ instead. By 
Eigendecomposition of $X^tX = V \Lambda V^t$ we can take $P = V^t X^t 
\Lambda^{-1/2}$, which also diagonalizes the scatter matrix $S$ but is 
more efficient to compute~\cite{Jegou2012,Bishop2006}.

To accommodate whitening, we set the weights of our first reduction layer 
to $W_1 = V^t X^t \Lambda^{-1} \sqrt{n}$ and keep them fixed during training. 

\noindent \textbf{Bagging.}
Since our first unsupervised layers can be fixed, we can train ensemble models
and average their predictions very efficiently for bagging
purposes~\cite{Maclin1997,Zhou2002,Perronnin2015} by caching the output of the
unsupervised layers and reusing it in the subsequent models.

%% file: experiments.tex
% !TEX root = main.tex

\section{Experiments}\label{sec:experiments}

We first describe the datasets used in our experiments, then provide a detailed
analysis of the iDT-FV pipeline and our proposed improvements. Based on our
observations, we then perform an ablative analysis of our proposed hybrid
architecture. Finally, we study the transferability of our hybrid models, and
compare to the state of the art.

\subsection{Datasets}

We use five publicly available and commonly used datasets for action
recognition. We briefly describe these datasets and their evaluation protocols. 

The \textbf{Hollywood2} \cite{Marszalek2009} dataset contains 1,707 
videos extracted from 69 Hollywood movies, distributed over 12 
overlapping action classes. As one video can have multiple class labels, 
results are reported using the mean average precision (mAP).

The \textbf{HMDB-51} \cite{Kuehne2011} dataset contains 6,849 videos
distributed over 51 distinct action categories. Each class contains at least
101 videos and presents a high intra-class variability. The evaluation
protocol is the average accuracy over three fixed splits~\cite{Kuehne2011}. 

The \textbf{UCF-101} \cite{Soomro2012} dataset contains 13,320 video clips
distributed over 101 distinct classes. This is the same dataset used in the
THUMOS'13 challenge \cite{Jiang2013}. The performance is again measured as the
average accuracy on three fixed splits.

The \textbf{Olympics} \cite{Niebles2010} dataset contains 783 
videos of athletes performing 16 different sport actions, with 50 
sequences per class. Some actions include interactions with objects, 
such as \textit{Throwing}, \textit{Bowling}, and \textit{Weightlifting}. 
Following \cite{Lan2014,Wang2013}, we report mAP over the train/test 
split released with the dataset.

The \textbf{High-Five} (a.k.a. TVHI) \cite{Patron-Perez2010} dataset contains
300 videos from 23 different TV shows distributed over four different human
interactions and a negative (no-interaction) class. As in
\cite{Patron-Perez2010,Patron-Perez2012,Wanga,Gaidon2014}, we report mAP for
the positive classes (mAP+) using the train/test split provided by the dataset
authors.

\input{expsShallow}

\input{expsHybrid}

\input{expsTransfer}

\input{expsSOTA}

% TODO EV: can we have one or two sentences somewhere about implementation details, e.g. done in Keras, etc.?

%% file: expsShallow.tex
% !TEX root = experiments.tex

\subsection{Detailed Study of Dense Trajectory Baselines for Action Recognition}\label{sec:expsShallow}

Table~\ref{table:shallow_ablative} reports our results comparing the iDT
baseline (Section~\ref{ss:idt}), its improvements discussed in
Section~\ref{ss:idtricks}, and our proposed data augmentation strategy
(Section~\ref{ss:feataug}). 

\noindent \textbf{Reproducibility}.
We first note that there are multiple differences in the iDT pipelines used
across the literature. While \cite{Wang2013} applies RootSIFT only on HOG, HOF,
and MBH, in \cite{Lan2014} this normalization is also applied to the Traj
descriptor.  While \cite{Wang2013} includes Traj in their pipeline,
\cite{Wanga} omits it.
Additionally, person bounding boxes are used to ignore human motions when doing
camera motion compensation in \cite{Wanga}, but are not publicly available for
all datasets.
Therefore, we reimplemented the main baselines and compare our results to the
officially published ones. As shown in Table~\ref{table:shallow_ablative}, we
successfully reproduce the original iDT results from \cite{Wang2013} and
\cite{Wang}, as well as the MIFS results of \cite{Lan2014}.

\noindent \textbf{Improvements of iDT.}
Table~\ref{table:shallow_ablative} shows that double-normalization 
(DN) alone improves performance over iDT on most datasets without the 
help of STA. We show that STA gives comparable results to SFV+STP, as 
hypothesized in section \ref{ss:idtricks}. Given that STA and DN are both 
beneficial for performance, we combine them with our own method. 

\noindent \textbf{Data Augmentation by Feature Stacking (DAFS).}
Although more sophisticated transformations can be used, we found that
combining a limited number of simple transformations already allows to
significantly improve the iDT-based methods in conjunction with the
aforementioned improvements, as shown in the ``iDT+STA+DAFS+DN'' line of
Table~\ref{table:shallow_ablative}.
In practice, we generate on-the-fly 7 different versions for each video,
considering the possible combinations of frame-skipping up to level 3 and
horizontal flipping.
%
% As an implementation detail, we use FFmpeg to generate those versions
% on-the-fly before extracting their feature matrices using iDT.
%
% After extraction, we stack those matrices together. Our results are shown in
% the last row of Table~\ref{table:shallow_ablative}. 
%
\noindent \textbf{Fine tuned and non-linear SVMs.} Attempting to
improve our best results, we also performed experiments both fine-tuning $C$
and also using a Gaussian kernel while fine-tuning $\gamma$. However, we
found that those two sets of experiments did not lead to significant improvements.
As DAFS already brings results competitive with the current state of the art, we
set those results with fixed $C$ as our current shallow baseline (FV-SVM). We will
now incorporate those techniques in the first unsupervised layers of our hybrid models. 

%
%Being very high dimensional and given the small number of training videos,
%the FV encoding almost always results in training sets that are close to linearly separable
%(but that do not necessarily yield optimal generalization).
%
%This indicates that our hybrid model improves performance not just because it is
%a non-linear classifier. Unlike one-vs-rest SVMs, our model can learn high-level
%discriminative representations shared across all the classes, which seems beneficial
%for classification.

%------- ABLATIVE ANALYSIS OF SHALLOW METHODS -------------
\begin{table}[t]
  \begin{threeparttable}
  \vspace{-0.4cm}
  \centering
    \caption{Analysis of iDT baselines and several improvements}
    \begin{tabular}{lccccccc}
        \toprule
        %                          re-checked                re-checked           re-checked    re-checked            re-checked
                                 & UCF-101                 & HMDB-51            & Hollywood2  & High-Five           & Olympics      \\
                                 & \%mAcc (s.d.)           & \%mAcc (s.d.)      & \%mAP       & \%mAP+ (s.d.)       & \%mAP         \\
    \midrule                                                                                                                        
     iDT \cite{Wang2013}         & 84.8 \cite{Wang}\tas\tda& 57.2               & 64.3        & -                   & 91.1          \\
     Our reproduction            & 85.0 (1.32)\tas\tda     & 57.0 (0.78)        & 64.2        & 67.7 (1.90)         & 88.6          \\
    \midrule                                                                                                                          
     iDT+SFV+STP \cite{Wanga}    & 85.7\tas\tda            & 60.1\tas           & 66.8\tas    & 68.1\tas\tda        & 90.4\tas      \\
     Our reproduction            & 85.4 (1.27)\tas\tda     & 59.3 (0.80)\tas    & 67.1\tas    & 67.8 (3.78)\tas\tda & 88.3\tas      \\
    \midrule                                                                                                                          
     iDT+STA+DN \cite{Lan2014}   & 87.3                    & 62.1               & 67.0        & -                   & 89.8          \\
     Our reproduction            & 87.3 (0.96)\tda         & 61.7 (0.90)        & 66.8        & 70.4 (1.63)         & 90.7          \\
    \midrule                                                                                                                          
     iDT+STA+MIFS+DN \cite{Lan2014} & 89.1                    & 65.1               & 68.0        & -                   & 91.4          \\
     Our reproduction            & 89.2 (1.03)\tda         & 65.4 (0.46)        & 67.1        & 70.3 (1.84)         & 91.1          \\
    \midrule                                                                                                                        
     iDT+DN                      & 86.3 (0.95)\tda         & 59.1 (0.45)        & 65.7        & 67.5 (2.27)         & 89.5          \\
     iDT+STA                     & 86.0 (1.14)\tda         & 60.3 (1.32)        & 66.8        & 70.4 (1.96)         & 88.2          \\
%     iDT+HFDA \cite{Fernando2015}& 86.4 (1.39)\tda         & 59.9 (0.96)        & 65.7        & 69.6 (1.58)         & 93.4          \\
%     iDT+HFFS                    & 86.7 (1.19)\tda         & 60.3 (0.46)        & 66.5        & 69.4 (1.90)         & 93.8          \\
%     iDT+DAFS                    & 88.6 (1.06)\tda         & 65.2 (0.99)        & 67.1        & 70.3 (1.93)         & 92.3          \\
%     iDT+HFFS+DN              & 88.3 (1.23)\tda         & 62.9 (0.47)        & 67.5        & 69.2 (2.49)         & 93.2          \\
%     iDT+STA+HFFS+DN          & 89.1 (1.21)\tda         & 65.2 (0.62)        & 68.8        & \textbf{71.1 (3.11)}& \textbf{93.3} \\
     \textbf{iDT+STA+DAFS+DN}    & \textbf{90.6 (0.91)\tda}&\textbf{67.8 (0.22)}&\textbf{69.1}& \textbf{71.0 (2.46)}& \textbf{92.8} \\
    \bottomrule
    \end{tabular}
    \label{table:shallow_ablative}
    \begin{tablenotes}
      \scriptsize
      \item iDT: Improved Dense Trajectories; SFV: Spatial Fisher Vector; STP: Spatio-Temporal Pyramids; STA: Spatio-Temporal Augmentation; MIFS: Multi-skIp Feature Stacking; DN: Double-Normalization; DAFS: Data Augmentation Feature Stacking; \tas without Trajectory descriptor; \tda without Human Detector.
    \end{tablenotes}
  \end{threeparttable}
  \vspace{-0.5cm}
\end{table}

%% file: expsHybrid.tex
% !TEX root = experiments.tex

\subsection{Analysis of Hybrid Models}\label{sec:expsHybrid}

\afterpage{
\begin{table}[t]
  \vspace{-0.4cm}
  \centering
  \small
    \caption{Top-5 best performing hybrid architectures with consistent improvements}
    \begin{tabular}{ccc|ccccc|c}
    \toprule
              &             &            &     UCF-101 & HMDB-51     & Hollywood2  & High-Five    & Olympics    &  Relative       \\
     Depth    &  Width      &   Batch    &    \%mAcc   & \%mAcc      & \%mAP       & \%mAP+      & \%mAP       &  Improv.       \\
    \midrule                                                                          
    \textbf{2}&\textbf{4096}&\textbf{128}&        91.6 &        68.1 &        72.6 &\textbf{73.1}&\textbf{95.3}&\textbf{2.46\%} \\
            2 &        4096 &        256 &        91.6 &        67.8 &        72.5 &        72.9 &        95.3 &        2.27\%  \\
            2 &        2048 &        128 &        91.5 &        68.0 &        72.7 &        72.7 &        94.8 &        2.21\%  \\
            2 &        2048 &        256 &        91.4 &        67.9 &        72.7 &        72.5 &        95.0 &        2.18\%  \\
            2 &         512 &        128 &        91.0 &        67.4 &\textbf{73.0}&        72.4 &        95.3 &        2.05\%  \\
    \midrule                                                                            
            1 &           - &          - & \textbf{91.9} & \textbf{68.5}&        70.4 &        71.9 &        93.5 &        1.28\%  \\
    \midrule                                                                            
     \multicolumn{3}{l|}{Best FV-SVM (\cf                                                 
     Tab.~\ref{table:shallow_ablative})}  &        90.6 &        67.8 &        69.1 &        71.0 &        92.8 &        0.00\%  \\
    \bottomrule
    \end{tabular}
  \label{table:hybrid_ablative}
  \vspace{-3mm}
\end{table}
%}
%\afterpage{
\begin{figure}[t]
    \begin{subfigure}{0.49\textwidth}
      \centering
      \includegraphics[width=1.1\linewidth]{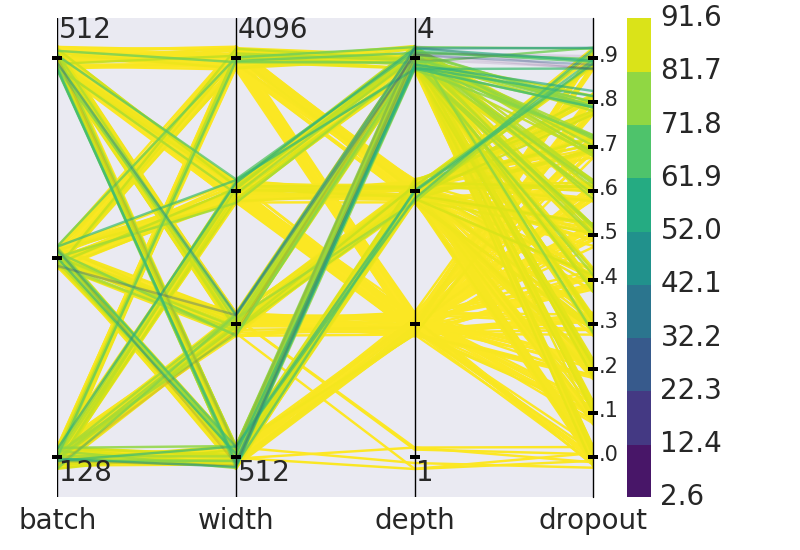}
	  \vspace{-6mm}
      \caption{\scriptsize UCF-101}
      \label{fig:sfig0}
    \end{subfigure}
%    \begin{subfigure}{.5\textwidth}
%      \centering
%      \includegraphics[width=1.1\linewidth]{img/pc/hi5.png}
%      \caption{High-Five}
%      \label{fig:sfig1}
%    \end{subfigure}
%    \begin{subfigure}{.5\textwidth}
%      \centering
%      \includegraphics[width=1.1\linewidth]{img/pc/hdb.png}
%      \caption{HMDB-51}
%      \label{fig:sfig2}
%    \end{subfigure}
%    \begin{subfigure}{.5\textwidth}
%      \centering
%      \includegraphics[width=1.1\linewidth]{img/pc/oly.png}
%      \caption{Olympics Sports}
%      \label{fig:sfig3}
%    \end{subfigure}
%    \begin{subfigure}{.5\textwidth}
%      \centering
%      \includegraphics[width=1.1\linewidth]{img/pc/hw2.png}
%      \caption{Hollywood2}
%      \label{fig:sfig4}
%    \end{subfigure}
    \begin{subfigure}{.49\textwidth}
      \centering
      \includegraphics[width=1.1\linewidth]{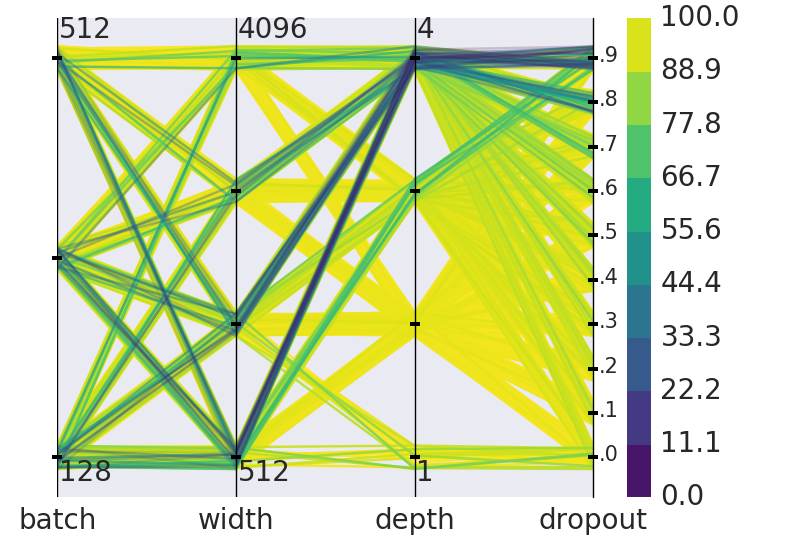}
	  \vspace{-6mm}
      \caption{\scriptsize All datasets (normalized)}
      \label{fig:sfig5}
    \end{subfigure}
    \begin{subfigure}{1.0\textwidth}
   	  \vspace{1mm}
      \small
      Each line represents one combination of parameters and color indicates performance
      of our hybrid architectures with unsupervised dimensionality reduction. Depth 2 correlates
      with high-performing architectures, whereas a small width and a large depth is suboptimal.
    \end{subfigure}
    \vspace{-2mm}
    \caption{Parallel coordinates plots showing the impact of multiple parameters.}
    \label{fig:hybrid_ablative}
    \vspace{-6mm}
\end{figure}
}

In this section, we start from hybrid architectures with unsupervised
dimensionality reduction learned by PCA.
For UCF-101 (the largest dataset) we initialize $W_1$ with $r=4096$
dimensions, whereas for all other datasets we use the number of dimensions
responsible for $99\%$ of the variance (yielding less dimensions than training
samples). 

We study the interactions between four parameters that can influence the
performance of our hybrid models: the output dimension of the intermediate
fully connected layers (\emph{width}), the number of layers (\emph{depth}), the
dropout rate, and the mini-batch size of Adam (\emph{batch}).
We systematically evaluate all possible combinations and rank the architectures
by the average relative improvement \wrt the best FV-SVM model. 
Training all 480 combinations for one split of UCF-101 can
be accomplished in less than two days with a single Tesla K80 GPU.
We report the top results in~Table~\ref{table:hybrid_ablative} and visualize
all results using the parallel coordinates plot in
Figure~\ref{fig:hybrid_ablative}. Our observations are as follows.

\noindent \textbf{Unsupervised dimensionality reduction.}
Performing dimensionality reduction using the weight matrix from PCA is
beneficial for all datasets, and using this layer alone, achieves 1.28\%
average improvement (Table \ref{table:hybrid_ablative}, depth 1) upon our
best SVM baseline.
% This improvement is in line with~\cite{Oruganti2016}.
% maybe cite earlier if really needed

\noindent \textbf{Width.}
We consider networks with fully connected layers of size 512, 1024, 2048, and
4096. We find that a large width (4096) gives the best results in 4 of 5
datasets.

\noindent \textbf{Depth.}
We consider hybrid architectures with depth between 1 and 4. Most
well-performing models have depth 2 as shown in
Figure~\ref{fig:hybrid_ablative}, but one layer is enough for the big datasets.

\noindent \textbf{Dropout rate.}
We consider dropout rates from 0 to 0.9. We find dropout to be dependent of
both architecture and dataset. A high dropout rate significantly impairs
classification results when combined with a small width and a large depth.

\noindent \textbf{Mini-batch size.}
We consider mini-batch sizes of 128, 256, and 512. We find lower batch sizes
to bring best results, with 128 being the more consistent across all datasets.
We observed that large batch sizes were detrimental to networks with a small
width.

\noindent \textbf{Best configuration with unsupervised dimensionality reduction.}
We find the following parameters to work the best: small batch sizes, a large
width, moderate depth, and dataset-dependent dropout rates. The most consistent
improvements across datasets are with a network with batch-size 128, width
4096, and depth 2.

\noindent \textbf{Supervised dimensionality reduction.}
Our previous findings indicate that the dimensionality reduction layer can have
a large influence on the overall classification results.
%(\cf~Table~\ref{table:hybrid_ablative}, depth 1).
%
Therefore, we investigate whether a \emph{supervised} dimensionality reduction
layer trained \emph{mid-to-end} with the rest of the architecture could improve
results further.
Due to memory limitations imposed by the higher number of weights to be learned
between our 116K-dimensional input FV representation and the intermediate
fully-connected layers, we decrease the maximum network width to 1024.
In spite of this limitation, our results in Table~\ref{tab:hybrid_sdr} show
that much smaller hybrid architectures with supervised dimensionality reduction
improve (on the larger UCF-101 and HMDB-51 datasets) or maintain (on the other
smaller datasets) recognition performance.
% Moreover, the depth of the network could also be decreased from two
% layers in the unsupervised dimensionality-reduction case to a single layer in
% the supervised case. This is expected, as by our previous observation a large
% number of layers does not match well with smaller number of neurons. % TODO

\begin{table}[t]
  \vspace{-0.4cm}
  \centering
    \caption{Supervised dimensionality reduction hybrid architecture evaluation}
    %\scriptsize
    \setlength\tabcolsep{3.0pt} % default value: 6pt
    \begin{tabular}{ccc|cccccc}
    \toprule
                  &            &            & UCF-101             & HMDB-51            & Hollywood2        & High-Five          & Olympics        \\
         Depth    &  Width     &   Batch    & \%mAcc  (s.d.)      & \%mAcc  (s.d.)     & \%mAP             & \%mAP+ (s.d.)      & \%mAP           \\    
    \midrule                                    
    1             & 1024       & 128        & 92.3 (0.77)         &\textbf{69.4 (0.16)}& 72.5      & 71.8 (1.37)        & 95.2            \\
    1    &        512 &        128 &\textbf{92.3 (0.70)} & 69.2 (0.09)        & 72.2              &        72.2 (1.14) & 95.2   \\
    2             & 1024       & 128        & 91.9 (0.78)         & 68.8 (0.46)        & 71.8              & 72.0 (1.03)        & 94.8            \\
    2             & 512        & 128        & 92.1 (0.68)         & 69.1 (0.36)        & 70.8              & 71.9 (2.22)        & 94.2            \\
    \midrule                                                                            
    \multicolumn{3}{l|}{Best unsup. (\cf Tab.~\ref{table:hybrid_ablative})}  
                  &  91.9 & 68.5 & \textbf{73.0} & \textbf{73.1} & \textbf{95.3} \\        
    \bottomrule
    \end{tabular}
  \label{tab:hybrid_sdr}
  \vspace{-0.5cm}
\end{table}

\noindent \textbf{Comparison to hybrid models for image recognition.} Our experimental
conclusions and optimal model differ from \cite{Perronnin2015}, both on unsupervised and supervised 
learning details (\eg dropout rate, batch size, learning algorithm), and in the usefulness of a supervised
dimensionality reduction layer trained mid-to-end (not explored in \cite{Perronnin2015}). 
%

%% file: expsTransfer.tex
% !TEX root = experiments.tex

\subsection{Transferability of Hybrid Models}\label{ss:trans}

In this section, we study whether the first layers of our architecture can be
transferred across datasets. As a reference point, we use the first split of
UCF-101 to create a base model and transfer elements from it to other
datasets. We chose UCF-101 for the following reasons: it is the largest
dataset, has the largest diversity in number of actions, and contains multiple
categories of actions, including human-object interaction, human-human
interaction, body-motion interaction, and practicing sports.

\noindent \textbf{Unsupervised representation layers.}
We start by replacing the dataset-specific GMMs with the GMMs from the base
model. Our results in the second row of Table~\ref{table:trans_hybrid} show
that the transferred GMMs give similar performance to the ones using
dataset-specific GMMs. This, therefore, greatly simplifies the task of learning
a new model for a new dataset. We keep the transferred GMMs fixed in the next
experiments.

\noindent \textbf{Unsupervised dimensionality reduction layer.}
Instead of configuring the unsupervised dimensionality reduction layer with
weights from the PCA learned on its own dataset, we configure it with the
weights learned in UCF-101. Our results are in the third row of Table
\ref{table:trans_hybrid}.
This time we observe a different behavior: for Hollywood2 and HMDB-51, the best
models were found without transfer, whereas for Olympics it did not have any
measurable impact. However, transferring PCA weights brings significant
improvement in High-Five. One of the reasons for this improvement is the
evidently smaller training set size of High-Five (150 samples) in contrast to
other datasets. The fact that the improvement becomes less visible as the
number of samples in each dataset increases (before eventually degrading
performance) indicates there is a threshold below which transferring starts to
be beneficial (around a few hundred training videos). 

\noindent \textbf{Supervised layers after unsupervised reduction.}
We also study the transferability of further layers in our architecture, after
the unsupervised dimensionality reduction transfer. We take the base model
learned in the first split of UCF-101, remove its last classification layer,
re-insert a classification layer with the same number of classes as the target
dataset, and fine-tune this new model in the target dataset, using an order of
magnitude lower learning rate. The results can be seen in the last row of
Table~\ref{table:trans_hybrid}.  The same behavior is observed for HMDB-51 and
Hollywood2. However, we notice a decrease in performance for High-Five and a
performance increase for Olympics. We attribute this to the presence of many
sports-related classes in UCF-101.

\noindent \textbf{Mid-to-end reduction and supervised layers.}
Finally, we study whether the architecture with supervised dimensionality
reduction layer transfers across datasets, as we did for the unsupervised
layers. We again replace the last classification layer from the corresponding
model learned on the first split of UCF-101, and fine-tune the whole
architecture on the target dataset.
Our results in the second and third rows of Table \ref{table:trans_sdr} show
that transferring this architecture brings improvements for Olympics and
HMDB-51, but performs worse than transferring unsupervised layers only on
High-Five.

\begin{table}[t]
\vspace{-4mm}
    \centering
    %\scriptsize
    \caption{Transferability experiments involving unsupervised dimensionality reduction}
    \setlength\tabcolsep{2.3pt} % default value: 6pt
    \begin{tabular}{ccc|cccc}
        \toprule
        Representation & Reduction    & Supervised & HMDB-51             & Hollywood2        & High-Five           & Olympics         \\
        Layers         & Layer        & Layers     & \%mAcc (s.d.)       & \%mAP             & \%mAP+ (s.d.)       & \%mAP            \\
        \midrule                                                                                  
           own         & own          & own        & 68.0 (0.65)         & \textbf{72.6}     & 73.1 (1.01)         & 95.3             \\
           UCF         & own          & own        & \textbf{68.0 (0.40)}& 72.4              & 73.7 (1.76)         & 94.2             \\
           UCF         & UCF          & own        & 66.5 (0.88)         & 70.0              & \textbf{76.3 (0.96)}& 94.0             \\
           UCF         & UCF          & UCF        & 66.8 (0.36)         & 69.7              & 71.8 (0.12)         & \textbf{96.0}    \\
        \bottomrule
    \end{tabular}
    \label{table:trans_hybrid}
    \vspace{-2mm}
\end{table}

\begin{table}[t]
   \vspace{-4mm}
   \centering
   %\scriptsize
   \caption{Transferability experiments involving supervised dimensionality reduction}
    \setlength\tabcolsep{6.0pt} % default value: 6pt
    \begin{tabular}{cc|cccc}
        \toprule   
        Representation & Supervised & HMDB-51              & Hollywood2   & High-Five           & Olympics        \\
        Layers         & Layers     & \%mAcc (s.d.)        & \%mAP        & \%mAP+ (s.d.)       & \%mAP           \\
        \midrule       
        own            & own        & 69.2 (0.09)          & 72.2         & 72.2 (1.14)         & 95.2            \\
        UCF            & own        & 69.4 (0.16)          &\textbf{72.5} & 71.8 (1.37)         & 95.2            \\
        UCF            & UCF        & \textbf{69.6 (0.36)} & 72.2         &\textbf{73.2 (1.89)} &\textbf{96.3}    \\
        \bottomrule
    \end{tabular}
    \label{table:trans_sdr}
    \vspace{-6mm}
\end{table}

%% file: expsSOTA.tex
% !TEX root = experiments.tex

\subsection{Comparison to the State of the Art}

In this section, we compare our best models found previously to the state of the art. 

\noindent \textbf{Best models.}
For UCF-101, the most effective model leverages its large training set using
supervised dimensionality reduction (\cf Table \ref{tab:hybrid_sdr}).
For HMDB-51 and Olympics, the best models result from transferring the
supervised dimensionality reduction models from the related UCF-101 dataset
(\cf Table~\ref{table:trans_sdr}).
Due to its specificity, the best architecture for Hollywood2 is based on
unsupervised dimensionality reduction learned on its own data (\cf
Table~\ref{table:hybrid_ablative}), although there are similarly-performing
end-to-end transferred models (\cf Table~\ref{table:trans_sdr}).
For High-Five, the best model is obtained by transferring the unsupervised
dimensionality reduction models from UCF-101 (\cf Table
\ref{table:trans_hybrid}).

\noindent \textbf{Bagging.}
As it is standard practice~\cite{Perronnin2015}, we take the best models and
perform bagging with 8 models initialized with distinct random initializations.
This improves results by around one point on average, and our final results
are in Table~\ref{table:sota}.
 
\noindent \textbf{Discussion.}
In contrast to \cite{Perronnin2015}, our models outperform the state of the art, 
including methods trained on massive labeled datasets like ImageNet or Sports-1M, 
confirming both the excellent performance and the data efficiency of our approach.
%
%As shown in Figure~\ref{fig:sota_prc}, our method leads to substantial
%improvements for all datasets considered.
%
Table \ref{table:failure} illustrates some failure cases of our methods. Confusion matrices 
and precision-recall curves for all datasets are available in the supplementary material 
for fine-grained analysis.

\begin{figure}[b!]
    \centering
    \vspace{-0.7cm}
    \begin{subfigure}{0.8\textwidth}      
        \includegraphics[width=1.0\linewidth]{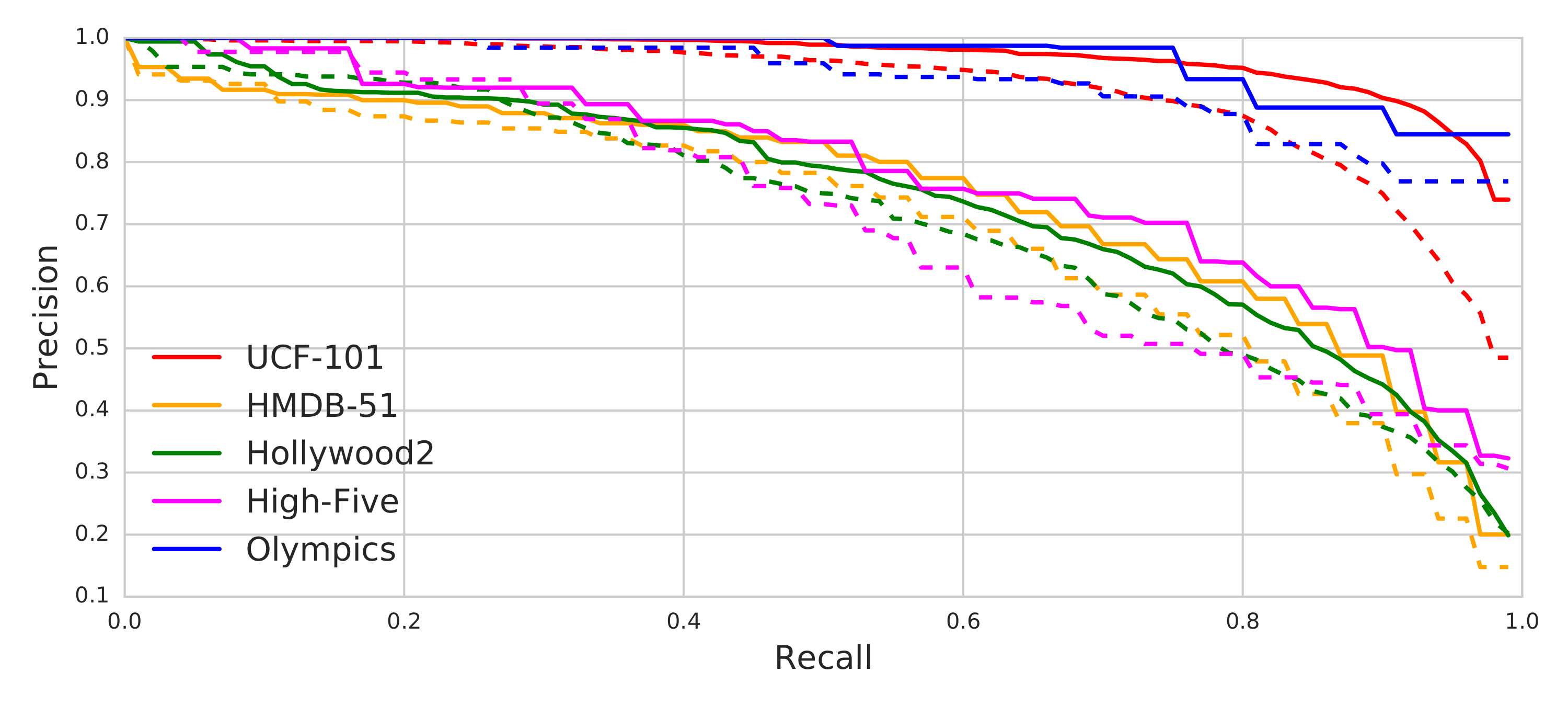}
    \end{subfigure}
    %\vspace{-0.5cm}
    \caption{Average Precision-Recall curves. Dashed: best FV-SVM, Full: best hybrid.
    }
    \label{fig:sota_prc}
\end{figure}

\begin{table}[!t]
  \vspace{-4mm}
  \centering
  %\scriptsize
  \begin{threeparttable}
    \caption{Comparison against the state of the art in action recognition}
    \scriptsize
    \setlength\tabcolsep{6.0pt}
    \setlength\aboverulesep{1.5pt}
    \setlength\belowrulesep{1.5pt}
    \begin{tabular}{p{3ex}lccccc}
        \toprule
                       &                                         & UCF-101             & HMDB-51             & Hollywood2                      & High-Five         & Olympics          \\
                       &  Method                                 & \%mAcc (s.d.)       & \%mAcc (s.d.)       & \%mAP                           & \%mAP+  (s.d.)      & \%mAP             \\
\midrule                                                                                                                                                                         
\tabr{8}{\textsc{Handcrafted}}                                                                                                                                                   
                       &iDT+FV\cite{Wang2013}                    & 84.8 \cite{Wang}    &  57.2               &    64.3                         & -                   & 91.1              \\ % 2013
                       &SDT-ATEP\cite{Gaidon2014}                &  -                  &  41.3               &    54.4                         & 62.4                & 85.5              \\ % 11-2013
                       &iDT+FM\cite{Peng2014a}                   & 87.9                &  61.1               &    -                            & -                   & -                 \\ % 05-2014 TODO: note: from arXiv
                       &RCS\cite{Hoai2014}                       &  -                  &  -                  &    73.6                         & 71.1                & -                 \\ % 11-2014                       
                       &iDT+SFV+STP\cite{Wanga}                  & 86.0                &  60.1               &    66.8                         & 69.4                & 90.4              \\ % 04-2015
                       &iDT+MIFS\cite{Lan2014}                   & 89.1                &  65.1               &    68.0                         & -                   & 91.4              \\ % 06-2015
                       &VideoDarwin\cite{Fernando2015}           &  -                  &  61.6               &    69.6                         & -                   & -                 \\ % 06-2015
                       &VideoDarwin+HF+iDT\cite{Fernando2015}    &  -                  &  63.7               & \textbf{73.7}                   & -                   & -                 \\ % 06-2015                       
\midrule                                                                                                                                                                         
\tabr{7}{\textsc{Deep-based}}                                                                                                                                                     
                       &2S-CNN\cite{Simonyan2014}\tup{IN}        & 88.0                &  59.4               &    -                            & -                   & -                 \\ % 12-2014
                       &2S-CNN+Pool\cite{Ng2015}\tup{IN}         & 88.2                &  -                  &    -                            & -                   & -                 \\ % 06-2015
                       &2S-CNN+LSTM\cite{Ng2015}\tup{IN}         & 88.6                &  -                  &    -                            & -                   & -                 \\ % 06-2015
                       &Objects+Motion(R*)\cite{Jain2015}\tup{IN}& 88.5                &  61.4               &    66.4                         & -                   & -                 \\ % 06-2015
                       &Comp-LSTM\cite{Srivastava2015}\tup{ID}   & 84.3                &  44.0               &    -                            & -                   & -                 \\ % 06-2015 (ICML2015)
                       &C3D+SVM\cite{Tran2014}\tup{S1M,ID}       & 85.2                &  -                  &    -                            & -                   & -                 \\ % 12-2015 (+ internal data)
                       &FSTCN\cite{Sun2015a}\tup{IN}             & 88.1                &  59.1               &    -                            & -                   & -                 \\ % 12-2015
%                      &Actions$\sim$Trans\cite{Wang2015}        & 92.0                &  62.0               &    -                            & -                   & -                 \\ % 07-2016
% TODO: INCLUDE OR NOT?&2S-Fusion\cite{Feichtenhofer2016}        & 93.5                &  69.2               &    -                            & -                   & -                 \\ % 07-2016 AFTER SUBMISSION DEADLINE                    
\midrule                                                                                                                                                                       
\tabr{6}{\textsc{Hybrid}}                                                                                                                                                        
                       &iDT+StackFV\cite{Peng2014b}              &  -                  &  66.8               &    -                            & -                   & -                 \\ % 09-2014  TODO:(shallow, deep or hybrid?)
                       &TDD\cite{Wang2015d}\tup{IN}              & 90.3                &  63.2               &    -                            & -                   & -                 \\ % 06-2015
                       &TDD+iDT\cite{Wang2015d}\tup{IN}          & 91.5                &  65.9               &    -                            & -                   & -                 \\ % 06-2015
                       &CNN-hid6\cite{Zha2015}\tup{S1M}          & 79.3                &  -                  &    -                            & -                   & -                 \\ % 09-2015 BMVC 2015
                       &CNN-hid6+iDT\cite{Zha2015}\tup{S1M}      & 89.6                &  -                  &    -                            & -                   & -                 \\ % 09-2015 TODO: look at the performance numbers for splits of UCF-101 in this paper: it is obvious that information from split 1 is leaking to the others (note how they are _much_ higher...)
                       &C3D+iDT+SVM\cite{Tran2014}\tup{S1M,ID}   & 90.4                &  -                  &    -                            & -                   & -                 \\ % 12-2015 (+ internal data)
\midrule                                                                                                                                                                      
                       & Best from state-of-the-art              &91.5 \cite{Wang2015d}&66.8 \cite{Peng2014b}&    \textbf{73.7} \cite{Fernando2015}     & 71.1 \cite{Hoai2014}&91.4 \cite{Lan2014}\\                                                 
\midrule
\midrule
                       & Our best FV+SVM                         & 90.6 (0.91)         &  67.8 (0.22)        &    69.1                         &        71.0 (2.46)  & 92.8             \\
                       % Our best hybrid (MLP)                   & 92.3 (0.77)         &  69.6 (0.36)        &    72.6                         &        76.3 (0.96)  & 96.3             \\
                       & Our best hybrid                         &\textbf{92.5 (0.73)} &\textbf{70.4 (0.97)} &    72.6                         &\textbf{76.7 (0.39)} &\textbf{96.7}     \\
                       % (with bagging)
\bottomrule
    \end{tabular}
    \begin{tablenotes}
      \scriptsize
	  \item Methods are organized by category (cf.\ Table~\ref{tab:related})
and sorted in chronological order in each block. Our hybrid models improve upon
the state of the art, and our handcrafted-shallow FV-SVM improves upon
competing end-to-end architectures relying on external data sources (IN: uses
ImageNet, S1M: uses Sports-1M, ID: uses private internal data).
% (\cite{Tran2014} pre-trains models on an internal I380K dataset; \cite{Srivastava2015} uses additional 300h of unrelated Youtube videos).
    \end{tablenotes}
\label{table:sota}
\end{threeparttable}
\vspace{-0.5cm}
\end{table}

\definecolor{gt}{HTML}{3EA055}
\definecolor{pr}{HTML}{C11B17}
\newcommand{\tgt}[1]{\fontsize{7}{5}\selectfont\color{gt}GT: #1}
\newcommand{\tpr}[1]{\fontsize{7}{5}\selectfont\color{pr}P: #1}
\newcommand{\specialcell}[2][c]{%
  \begin{tabular}[#1]{@{}c@{}}#2\end{tabular}}

\begin{table}[htbp]
  \centering
    \caption{Top-5 most confused classes for our best FV-SVM and Hybrid models}
    \begin{tabular}{cccccccc}
    \toprule
\multicolumn{2}{l}{\specialcell[c]{
                Top-5 \\ 
                Confusions}}           & \#1               & \#2                  & \#3                   & \#4                  & \#5                \\
    \midrule                                                                                                                     
    \tabr{6}{UCF-101}   &\tabrr{Hybrid}&\tabg{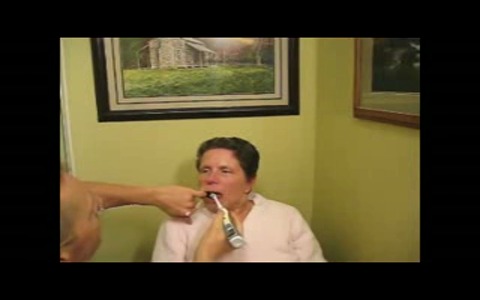} &\tabg{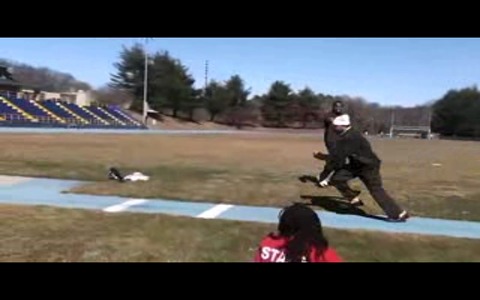}    &\tabg{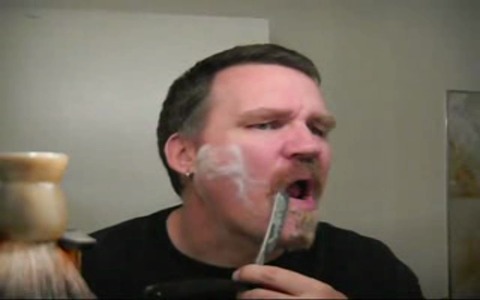}     &\tabg{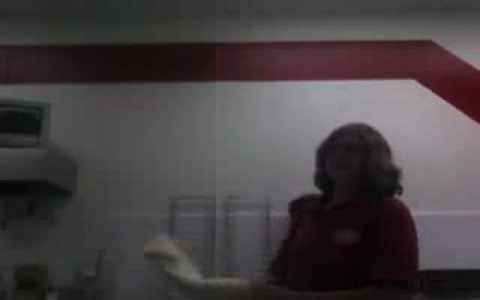}    &\tabg{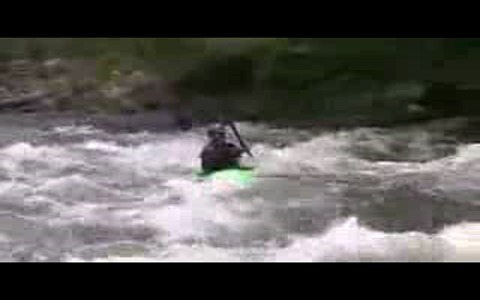}  \\[-0.7ex]
                        &              &\tpr{ShavingBeard} &\tpr{FrisbeeCatch}    &\tpr{ApplyLipstick}   &\tpr{Nunchucks}       &\tpr{Rafting}        \\[-0.7ex]
                        &              &\tgt{BrushingTeeth}&\tgt{LongJump}        &\tgt{ShavingBeard}    &\tgt{PizzaTossing}    &\tgt{Kayaking}       \\[0.6ex]
                        &\tabrr{FV-SVM}&\tabg{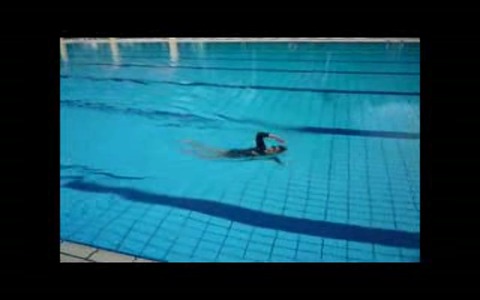} &\tabg{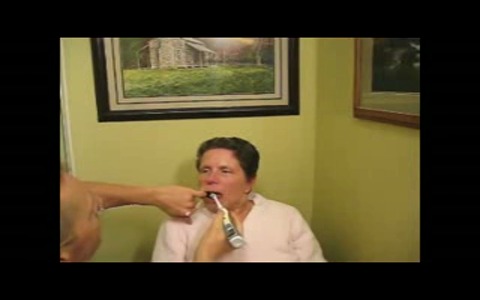}    &\tabg{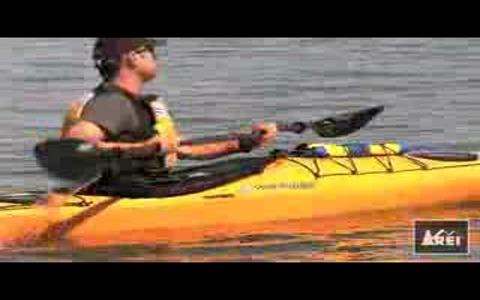}    &\tabg{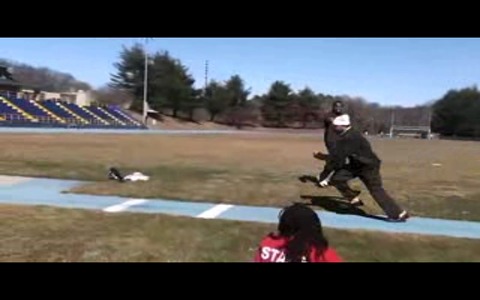}    &\tabg{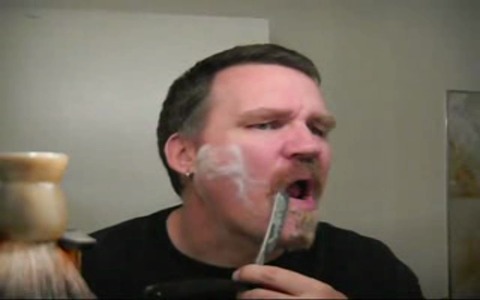}   \\[-0.7ex]
                        &              &\tpr{BreastStroke} &\tpr{ShavingBeard}    &\tpr{Rafting}         &\tpr{FrisbeeCatch}    &\tpr{ApplyLipstick}  \\[-0.7ex]
                        &              &\tgt{FrontCrawl}   &\tgt{BrushingTeeth}   &\tgt{Kayaking}        &\tgt{LongJump}        &\tgt{ShavingBeard}   \\[-0.5ex]
    \midrule                                                                                                                                          
    \tabr{6}{HMDB-51}   &\tabrr{Hybrid}&\tabg{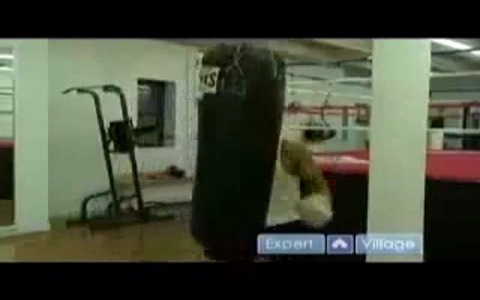} &\tabg{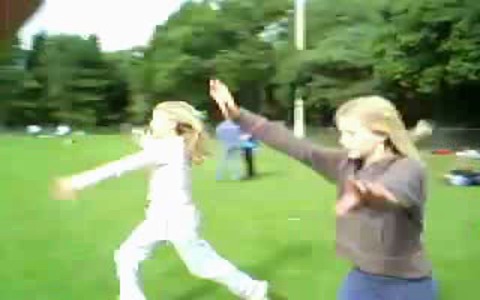}    &\tabg{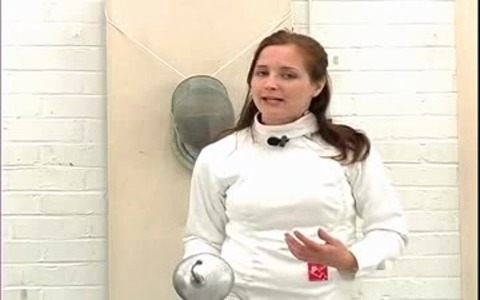}    &\tabg{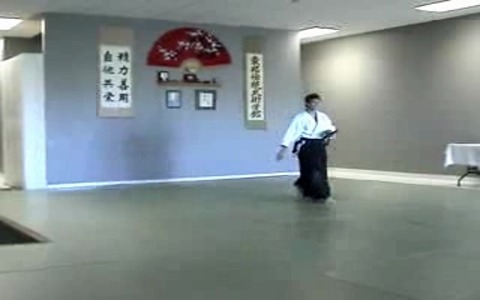}    &\tabg{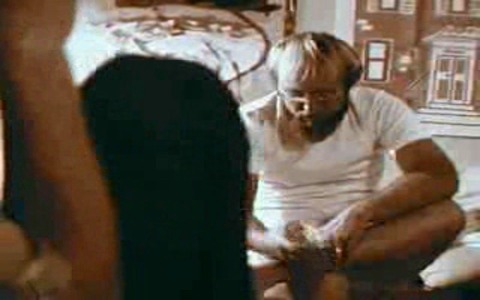}   \\[-0.7ex]
                        &              &\tpr{sword}        &\tpr{flic\_flac}      &\tpr{draw\_sword}     &\tpr{sword\_exercise} &\tpr{drink}          \\[-0.7ex]
                        &              &\tgt{punch}        &\tgt{cartwheel}       &\tgt{sword\_exercise} &\tgt{draw\_sword}     &\tgt{eat}            \\[0.6ex]
                        &\tabrr{FV-SVM}&\tabg{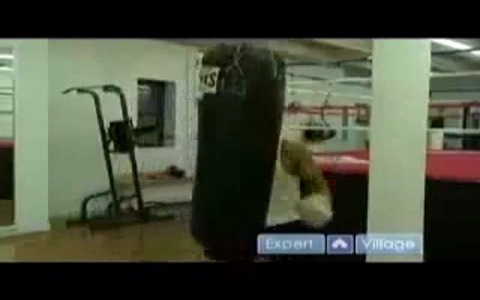} &\tabg{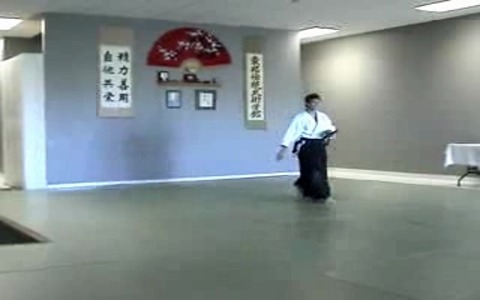}    &\tabg{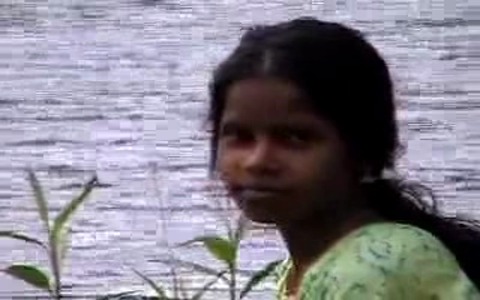}    &\tabg{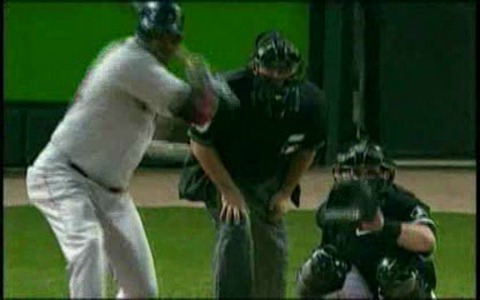}    &\tabg{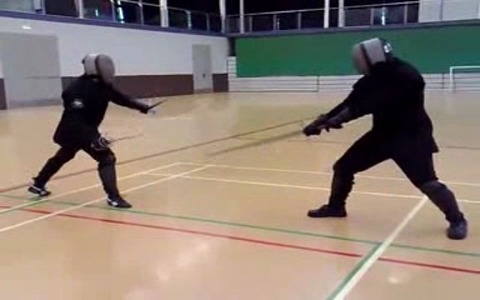}   \\[-0.7ex]
                        &              &\tpr{sword}        &\tpr{sword\_exercise} &\tpr{chew}            &\tpr{throw}           &\tpr{fencing}        \\[-0.7ex]
                        &              &\tgt{punch}        &\tgt{draw\_sword}     &\tgt{smile}           &\tgt{swing\_baseball} &\tgt{sword}          \\[-0.5ex]
    \midrule                                                                                                                                                                                                                                                                              
   \tabr{6}{High-Five} &\tabrr{Hybrid}&\tabg{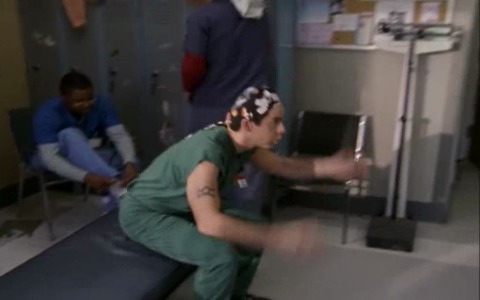} &\tabg{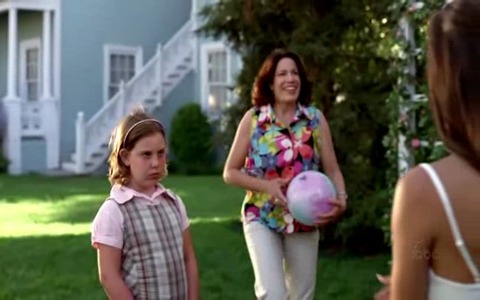}    &\tabg{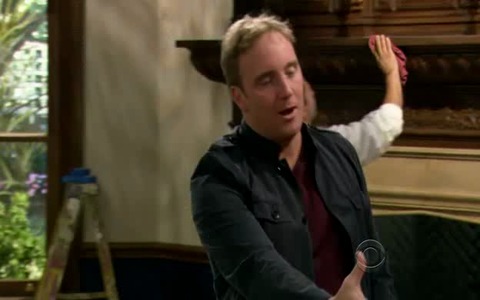}    &\tabg{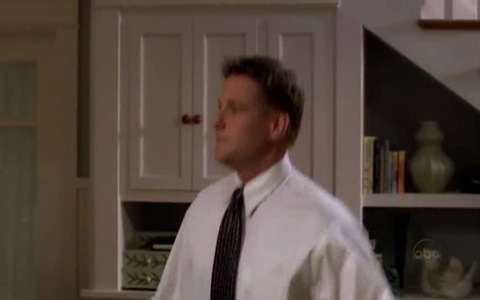}    &\tabg{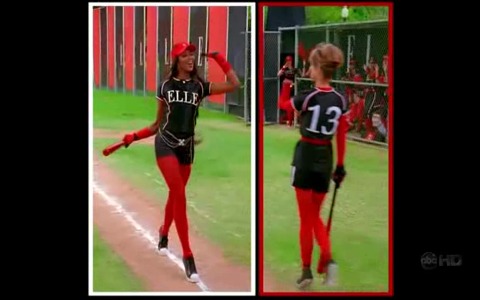}    \\[-0.7ex]
                       &              &\tpr{negative}     &\tpr{negative}        &\tpr{hug}             &\tpr{hug}             &\tpr{handShake}       \\[-0.7ex]
                       &              &\tgt{highFive}     &\tgt{handShake}       &\tgt{handShake}       &\tgt{kiss}            &\tgt{highFive}        \\[0.6ex]
                       &\tabrr{FV-SVM}&\tabg{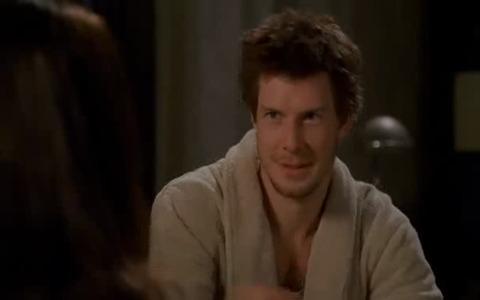} &\tabg{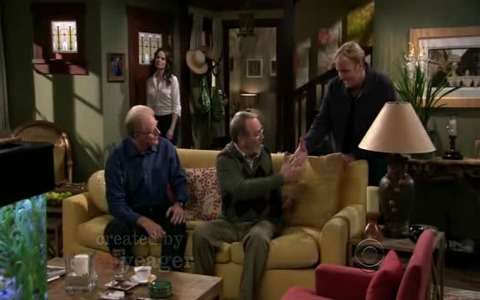}    &\tabg{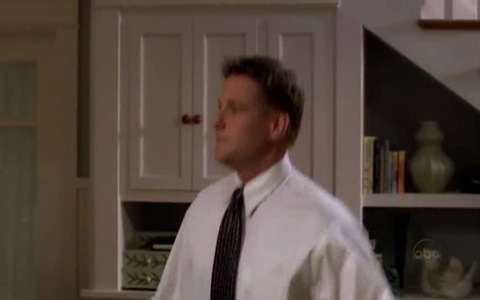}    &\tabg{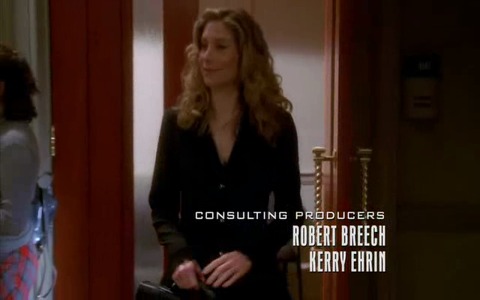}    &\tabg{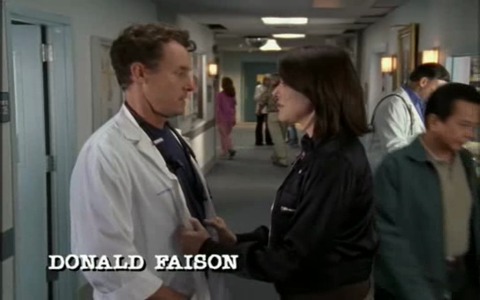}    \\[-0.7ex]
                       &              &\tpr{negative}     &\tpr{negative}        &\tpr{hug}             &\tpr{hug}             &\tpr{negative}        \\[-0.7ex]
                       &              &\tgt{highFive}     &\tgt{handShake}       &\tgt{kiss}            &\tgt{handShake}       &\tgt{kiss}            \\[-0.5ex]
   \midrule                                                                                                                                          
    \tabr{6}{Olympics}  &\tabrr{Hybrid}&\tabg{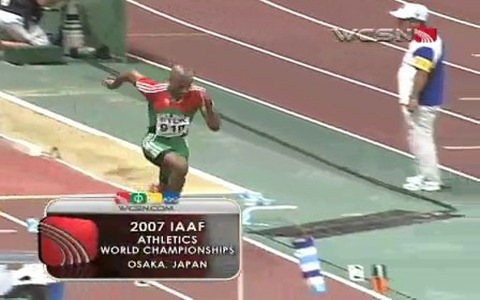} &\tabg{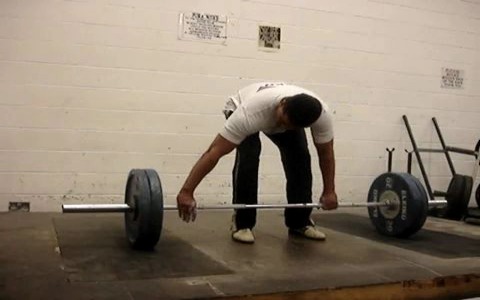}    &\tabg{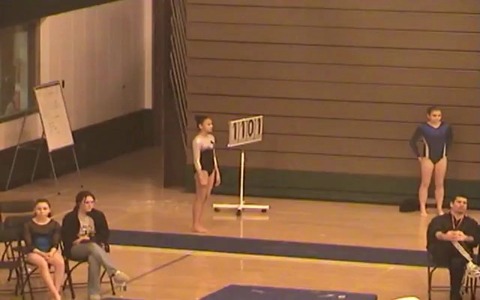}    &\tabg{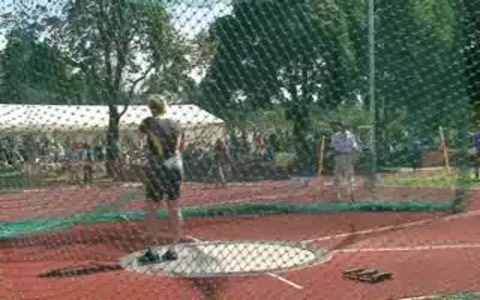}    &\tabg{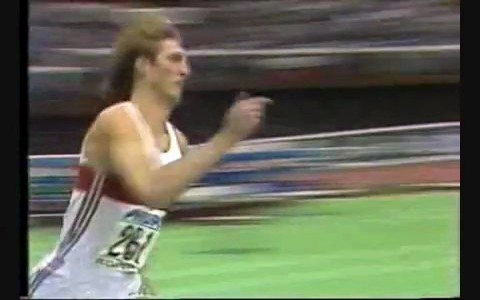}   \\[-0.7ex]
                        &              &\tpr{long\_jump}   &\tpr{clean\_and\_jerk}&\tpr{high\_jump}      &\tpr{discus\_throw}   &\tpr{vault}          \\[-0.7ex]
                        &              &\tgt{triple\_jump} &\tgt{snatch}          &\tgt{vault}           &\tgt{hammer\_throw}   &\tgt{high\_jump}     \\[0.6ex]
                        &\tabrr{FV-SVM}&\tabg{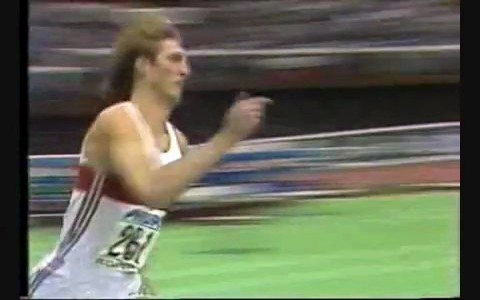} &\tabg{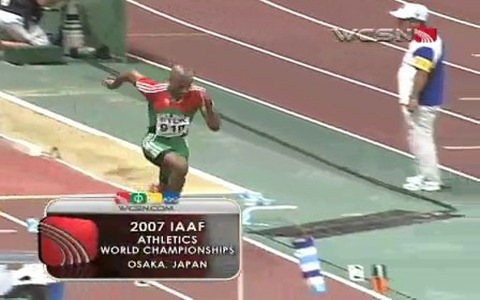}    &\tabg{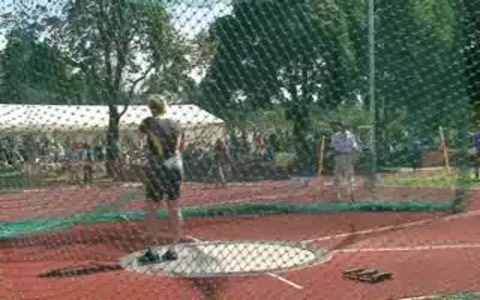}    &\tabg{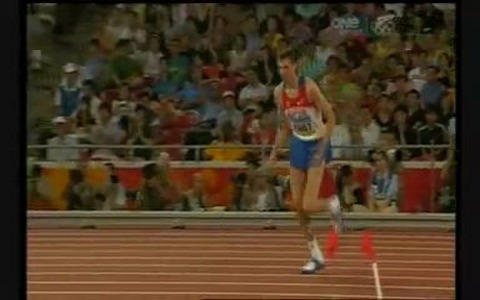}    &\tabg{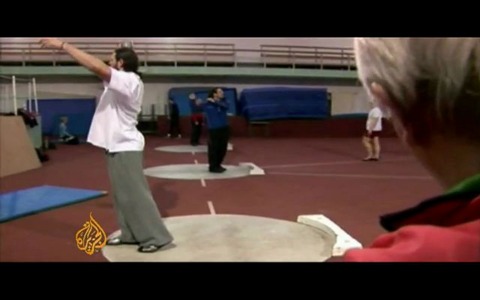}   \\[-0.7ex]
                        &              &\tpr{vault}        &\tpr{long\_jump}      &\tpr{discus\_throw}   &\tpr{pole\_vault}     &\tpr{bowling}        \\[-0.7ex]
                        &              &\tgt{high\_jump}   &\tgt{triple\_jump}    &\tgt{hammer\_throw}   &\tgt{high\_jump}      &\tgt{shot\_put}      \\[-0.5ex]
    \bottomrule
    \end{tabular}
\label{table:failure}
\end{table}

%% file: conclusion.tex
% !TEX root = main.tex

\section{Conclusion}\label{sec:conclusion}

We investigate hybrid architectures for action recognition, effectively
combining hand-crafted spatio-temporal features, unsupervised representation
learning based on the FV encoding, and deep neural networks.
In addition to paying attention to important details like normalization and
spatio-temporal structure, we integrate data augmentation at the feature level,
end-to-end supervised dimensionality reduction, and modern optimization and
regularization techniques.
We perform an extensive experimental analysis on a variety of datasets, showing
that our hybrid architecture yields data efficient, transferable models of small
size that yet outperform much more complex deep architectures trained
end-to-end on millions of images and videos.
We believe our results open interesting new perspectives to design even more
advanced hybrid models, \eg using recurrent neural networks, targeting better
accuracy, data efficiency, and transferability.

\noindent \textbf{Acknowledgements.}
Antonio M. Lopez is supported by the Spanish MICINN project TRA2014-57088-C2-1-R, 
and by the Secretaria d'Universitats i Recerca del Departament d'Economia i 
Coneixement de la Generalitat de Catalunya (2014-SGR-1506).

%% file: supp.tex
% !TEX root = main.tex

%\pagestyle{headings}
%\mainmatter

\title{\textls[-3]{Sympathy for the Details: Dense Trajectories and Hybrid Classification Architectures for Action Recognition}}

\subtitle{Supplementary material}

\titlerunning{Supplementary material for Sympathy for the Details: Dense Trajectories and ...
    %Architectures for Action Recognition
}

\authorrunning{C\'{e}sar de Souza, Adrien Gaidon, Eleonora Vig, Antonio L\'{o}pez}

\author{C\'{e}sar Roberto de Souza\textsuperscript{1,2}, Adrien 
    Gaidon\textsuperscript{1}, Eleonora Vig\textsuperscript{3}, 
    Antonio Manuel L\'{o}pez\textsuperscript{2}}

\institute{
    \textsuperscript{1}Computer Vision Group, Xerox Research Center Europe, Meylan, France \\
    \textsuperscript{2}Centre de Visi\'{o} per Computador, Universitat Aut\`{o}noma de Barcelona, Bellaterra, Spain \\
    \textsuperscript{3}German Aerospace Center, Wessling, Germany \\
    {\tt\small \{cesar.desouza, adrien.gaidon\}@xrce.xerox.com, eleonora.vig@dlr.de, antonio@cvc.uab.es}
}

\def\thefootnote{}
\maketitle

\input{abbreviations}
\section{Introduction}

This material provides additional details regarding our 
submission to ECCV16. It presents graphs and plots that helped guide our 
experiments and consolidate our findings, but that were tangential to the main 
objectives of the submission. In particular, we provide the complete parallel 
coordinates plots for all datasets considered (whereas the submission shows only 
UCF-101 and overall combined results) and the confusion matrices and 
precision-recall plots detailing the failure and improvement cases between our 
hybrid models and our Fisher Vector with Support Vector Machines (FV-SVM) 
baseline.

\section{Hybrid Unsupervised and Supervised Architectures}
\label{ss:best_archs}

\begin{figure}[p!]
    \begin{subfigure}{0.5\textwidth}
        \centering
        \includegraphics[width=1.1\linewidth]{img/pc/ucf.png}
        \vspace{-6mm}
        \caption{UCF-101}
        \label{fig:sfig0}
    \end{subfigure}
    \begin{subfigure}{.5\textwidth}
        \centering
        \includegraphics[width=1.1\linewidth]{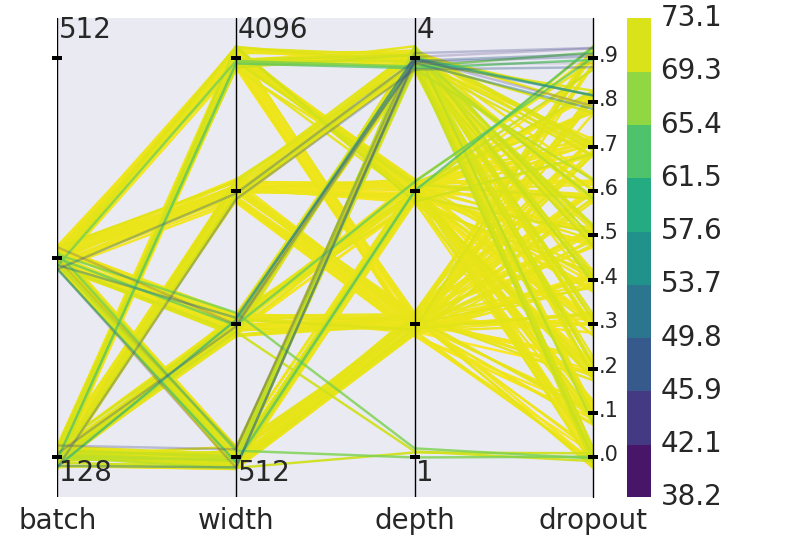}
        \vspace{-6mm}
        \caption{High-Five}
        \label{fig:sfig1}
    \end{subfigure}
    \begin{subfigure}{.5\textwidth}
        \centering
        \vspace{4mm}
        \includegraphics[width=1.1\linewidth]{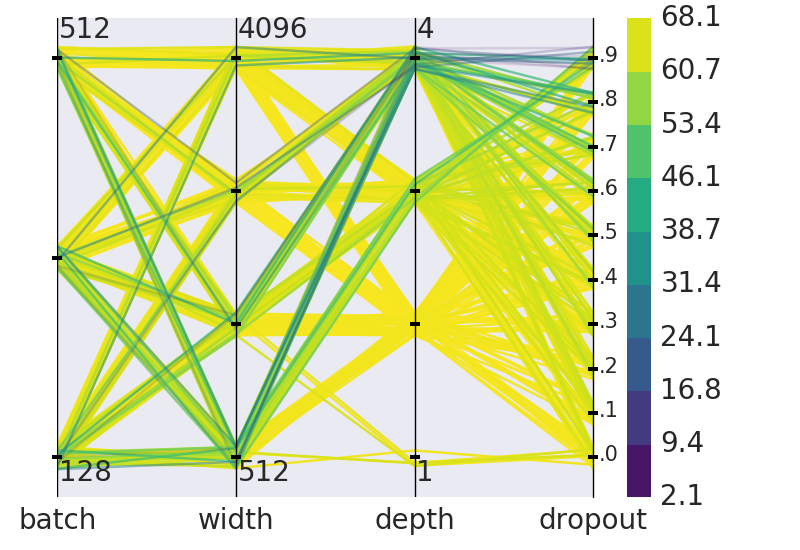}
        \vspace{-6mm}
        \caption{HMDB-51}
        \label{fig:sfig2}
    \end{subfigure}
    \begin{subfigure}{.5\textwidth}
        \centering
        \vspace{4mm}
        \includegraphics[width=1.1\linewidth]{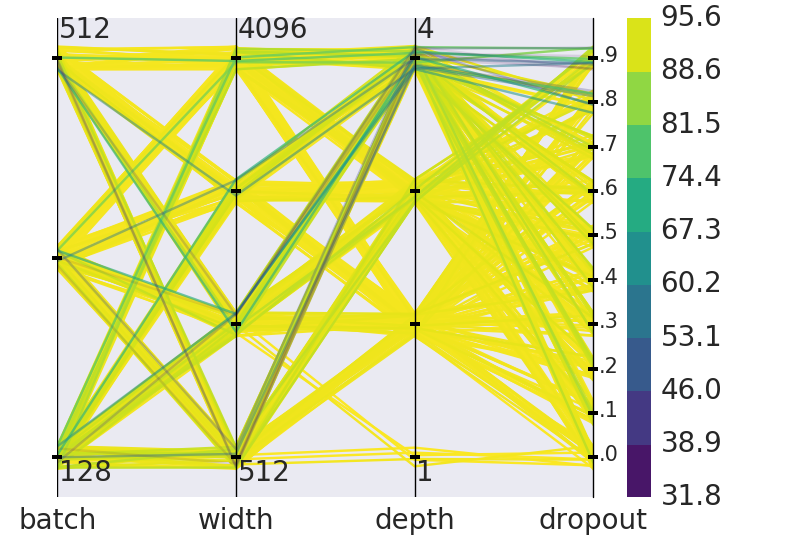}
        \vspace{-6mm}
        \caption{Olympic Sports}
        \label{fig:sfig3}
    \end{subfigure}
    \begin{subfigure}{.5\textwidth}
        \centering
        \vspace{4mm}
        \includegraphics[width=1.1\linewidth]{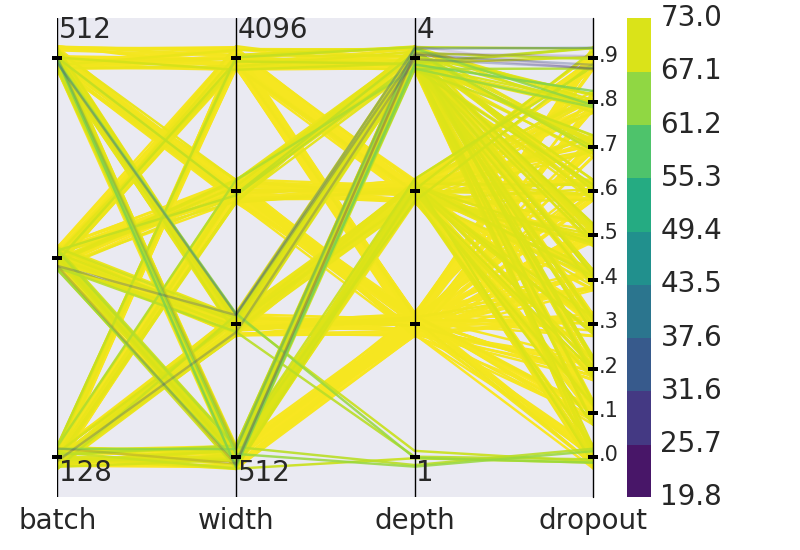}
        \vspace{-6mm}
        \caption{Hollywood2}
        \label{fig:sfig4}
    \end{subfigure}
    \begin{subfigure}{.5\textwidth}
        \centering
        \vspace{4mm}
        \includegraphics[width=1.1\linewidth]{img/pc/all.png}
        \vspace{-6mm}
        \caption{Combined (normalized)}
        \label{fig:sfig5}
    \end{subfigure}
    \caption{
        \small
        Parallel Coordinates plots showing the impact of multiple parameters on our
        hybrid architectures with unsupervised dimensionality reduction for each
        dataset. Each line represents one combination of parameters and colour
        indicates performance. Depth 2 correlates with high-performing architectures,
        whereas a small width and a large depth is suboptimal. The combined plot (f)
        has been generated by normalizing the performance in each dataset to the
        [0-100] interval and stacking the architectures tuples together (\cf
        explanatory text in Section 
        \ref{ss:best_archs}).}
    \label{fig:hybrid_ablative}
    \vspace{-6mm}
\end{figure}

We present per-dataset parallel coordinates plots showing all the explored
hybrid architectures with \emph{unsupervised} dimensionality reduction,
expanding Figure 2 of the submitted paper. The complete parallel coordinates
plots for all datasets can be seen in Figure \ref{fig:hybrid_ablative} and are
explained below.

\noindent \textbf{Explored architectures.}
Our parametric study, whose overall conclusion is described in Section 5.3 of
the main submission, evaluates \emph{480} architectures \emph{for each
dataset}, corresponding to three batch sizes $(128, 256, 512)$, four widths
$(512, 1024, 2048, 4096)$, four depths $(1, 2, 3, 4)$ and ten dropout rates
$(0.0, 0.1, ..., 0.9)$. 
We recall that by \emph{depth} we mean number of network layers, and by
\emph{width} we mean dimension of the intermediate fully connected layers (\cf
main text).

\noindent \textbf{Explanation of Parallel Coordinates (PC).}
PC plots are a visualization technique for displaying high-dimensional data in
2D. They can highlight meaningful multivariate patterns, especially when used
interactively~\cite{Few2006}.
In a PC plot, each data dimension is associated with a vertical line crossing
the $x$-axis. Plotting a single multi-dimensional data sample involves two
steps: (i) placing each dimension on its related vertical line, then (ii)
connecting these points with a line of the same color, thus creating a path
that crosses all the $y$-axes. Each dimension is normalized to the unit
interval before plotting.
In our case, each data point is a quadruplet $(batch, width, depth, dropout)$
coupled with a recognition performance number. Each hyper-parameter value has a
position along the $y$-axis corresponding to this hyper-parameter. We slightly
shift that position randomly for better readability.
The recognition performance of the path corresponding to a set of
hyper-parameters is encoded in the color of the line (brighter is better) and
its transparency (more solid is better).
% resulting in a combined plot with 2400 paths.

\noindent \textbf{Analysis.}
As mentioned in Section 5.3 of the main text, a clear pattern that appears in
the PC plots is the elevated brightness of depth 2 for all datasets.  Another
visible pattern is the relationship between lower widths and higher depths,
which is always marked by the presence of suboptimal architectures (dark
lines from width 512 and 1024 to depth 4). Other patterns associated with
suboptimal architectures appear between large batch sizes and a small width
(dark lines from batch 512 to width 512).
On the other hand, connections between large widths and depth 2 are always
clear in the plots for all datasets. This pattern supports the best
architecture found by our systematic ranking of the architectures discussed in
Section 5.3 of our submission: batch size of $128$, width of $4096$, and depth
$2$.
Regarding the dropout rate, the PC plots in Figure~\ref{fig:hybrid_ablative}
suggest that too high rates might be detrimental to the model, and the rate
should be tuned depending on the rest of the architecture and on the dataset.

\section{Failure and Success Cases per Dataset}

In this section we provide a detailed analysis for failure and success cases 
for all datasets considered, contrasting our best hybrid models against our 
strong FV-SVM baseline.  

The datasets used in our experiments can be divided according to two criteria:
the presence of overlapping classes (multi-label or multi-class) and
performance metric (mean Average Precision, mAP; or mean accuracy, mAcc).
In this section we present precision-recall curves for datasets whose
performance is computed using mAP (Hollywood2, High-Five, Olympics) and
confusion matrices for datasets which are multi-class (UCF-101, HMDB-51,
Olympics, High-Five).
We cluster the rows and columns of the confusion matrices by level of
confusion, which we obtain by ordering the rows via a 1D Locally Linear
Embedding (LLE) \cite{Roweis2000}. For the largest datasets, we show only the
top-25 confusion classes as determined by this embedding.
For the precision-recall curves, we show at every recall decile the quantized
cumulated precision segmented per class.

\subsection{UCF-101}

\begin{figure}[tbhp]
    \vspace{-1.5cm}
    \hspace{-0.5cm}
    \begin{subfigure}{0.575\textwidth}
        \centering
        \includegraphics[width=1.0\linewidth,
        trim={0 0 3cm 0},clip]{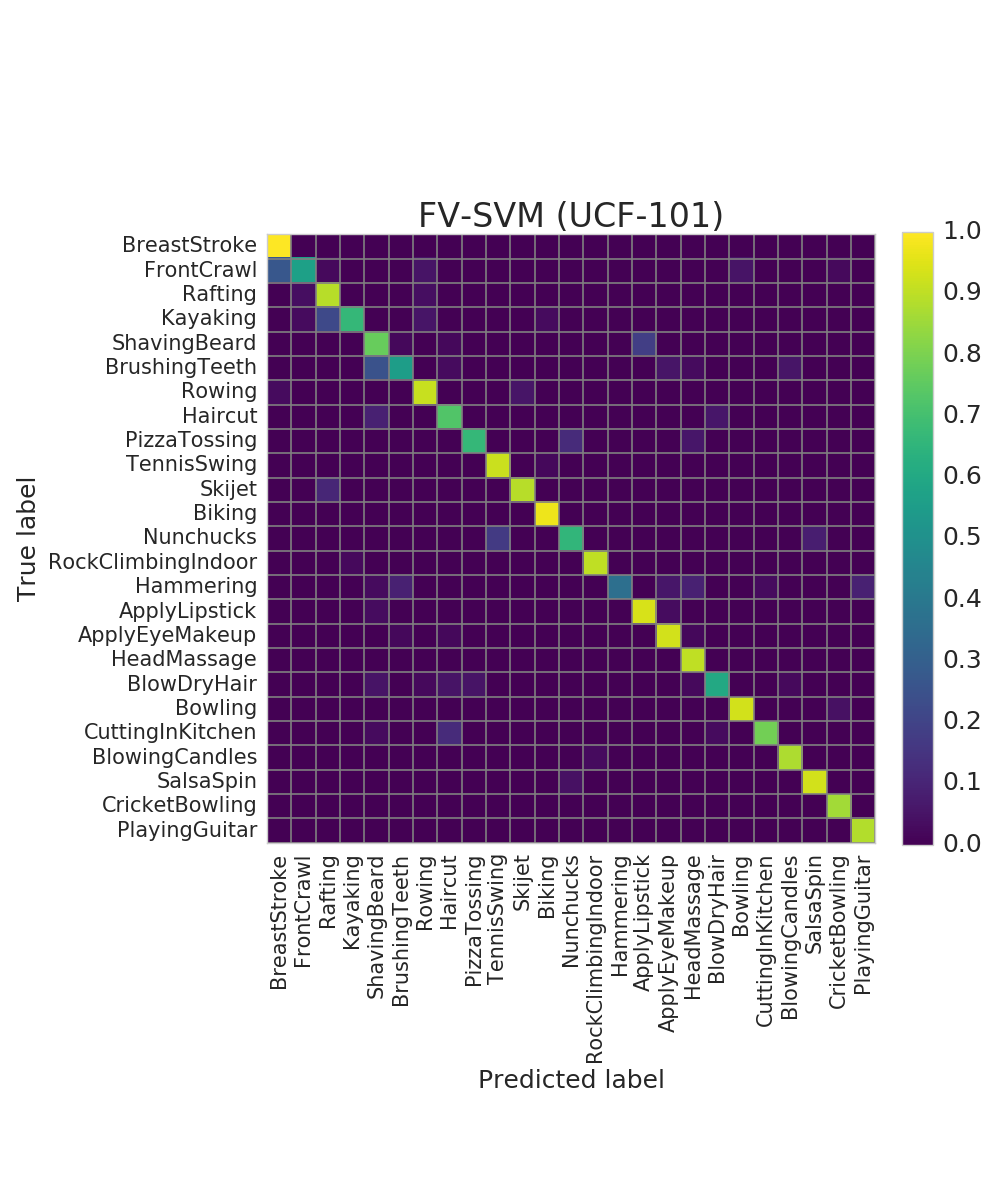}
        %\caption{FV-SVM}
    \end{subfigure}
    \begin{subfigure}{0.48\textwidth}
        \centering
        \includegraphics[width=1.0\linewidth,
        trim={6.6cm 0 0 0},clip]{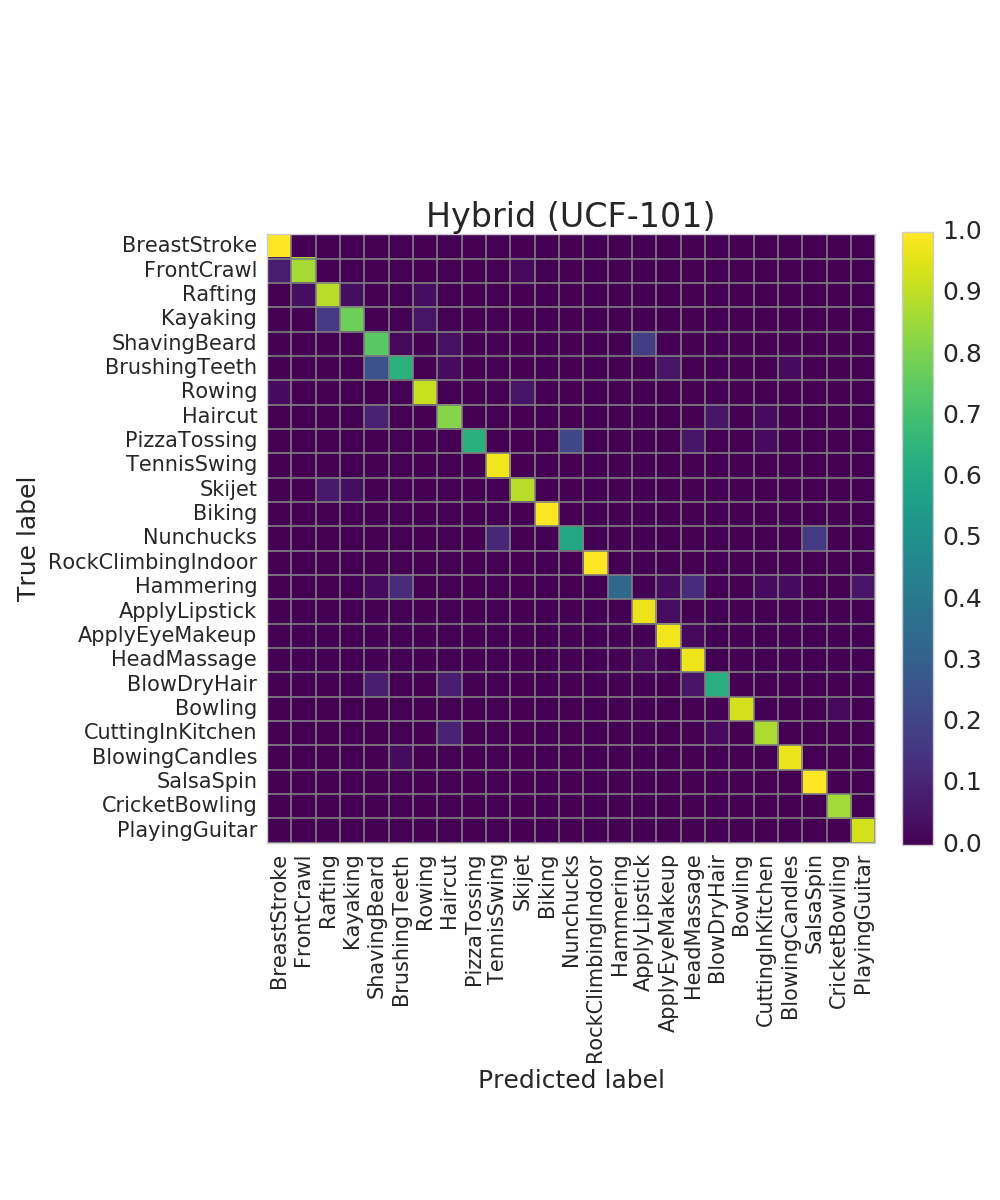}
        %\caption{Hybrid}
    \end{subfigure}
    %\vspace{-1cm}
    \caption{Top-25 most confused classes for UCF-101.}
    \label{fig:cm_ucf}
    \vspace{-0.5cm}
\end{figure}

UCF-101 is the largest dataset we investigate. Using split 1, we determine the
top 25 classes that are responsible for most of the confusion of our baseline
model (FV-SVM). Then, we show the evolution of those same classes in the
confusion matrix of our best hybrid model (Hybrid) in Figure~\ref{fig:cm_ucf}. 
Note that the differences are small overall, as both the FV-SVM and Hybrid
models have excellent recognition performance ($90.6\%$ and $92.5\%$ resp.,
\cf Table 8 in the main text).

The most confused classes are \textit{BreastStroke}, \textit{FrontCrawl},
\textit{Rafting}, \textit{Kayaking}, \textit{ShavingBeard} and \textit{Brushing
Teeth}. Inspecting both confusion matrices, we can see how the Hybrid model
solves most of the confusion between the \textit{BreastStroke} and
\textit{FrontCrawl} swimming actions. It also solves a visible confusion
between \textit{FrontCrawl} and \textit{Rowing}. Furthermore, it also improves
many other classes, such as \textit{ApplyLipstick}, \textit{ApplyEyeMakeUp}, and
\textit{HeadMassage}, but seems to reinforce certain mistakes such as the
confusion between \textit{Nunchucks} and \textit{SalsaSpin}.
Our results therefore suggest that our higher-capacity hybrid models can
improve on some fine-grained recognition tasks, but not all (\ie
\textit{Rafting} and \textit{Kayaking}), which confirms the previously known
good performance of FV and linear classifiers for fine-grained
tasks~\cite{Gosselin2014}.

\subsection{HMDB-51}

\begin{figure}[tbhp]
    \vspace{-1.5cm}
    \hspace{-0.5cm}
    \begin{subfigure}{0.55\textwidth}
        \centering
        \includegraphics[width=1.0\linewidth,
        trim={0 0 3cm 0},clip]{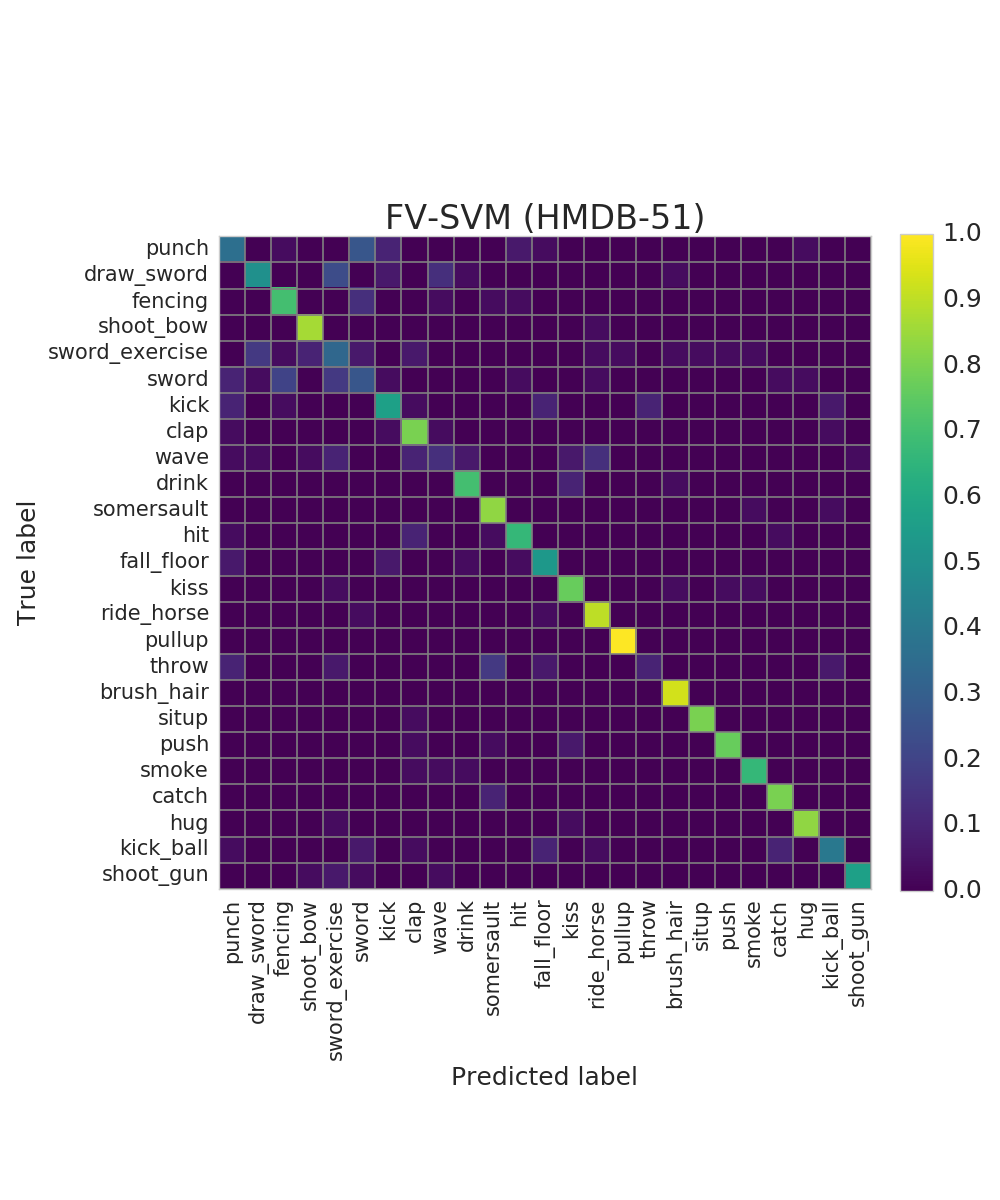}
    \end{subfigure}
    \begin{subfigure}{0.49\textwidth}
        \centering
        \includegraphics[width=1.0\linewidth,
        trim={5.5cm 0 0 0},clip]{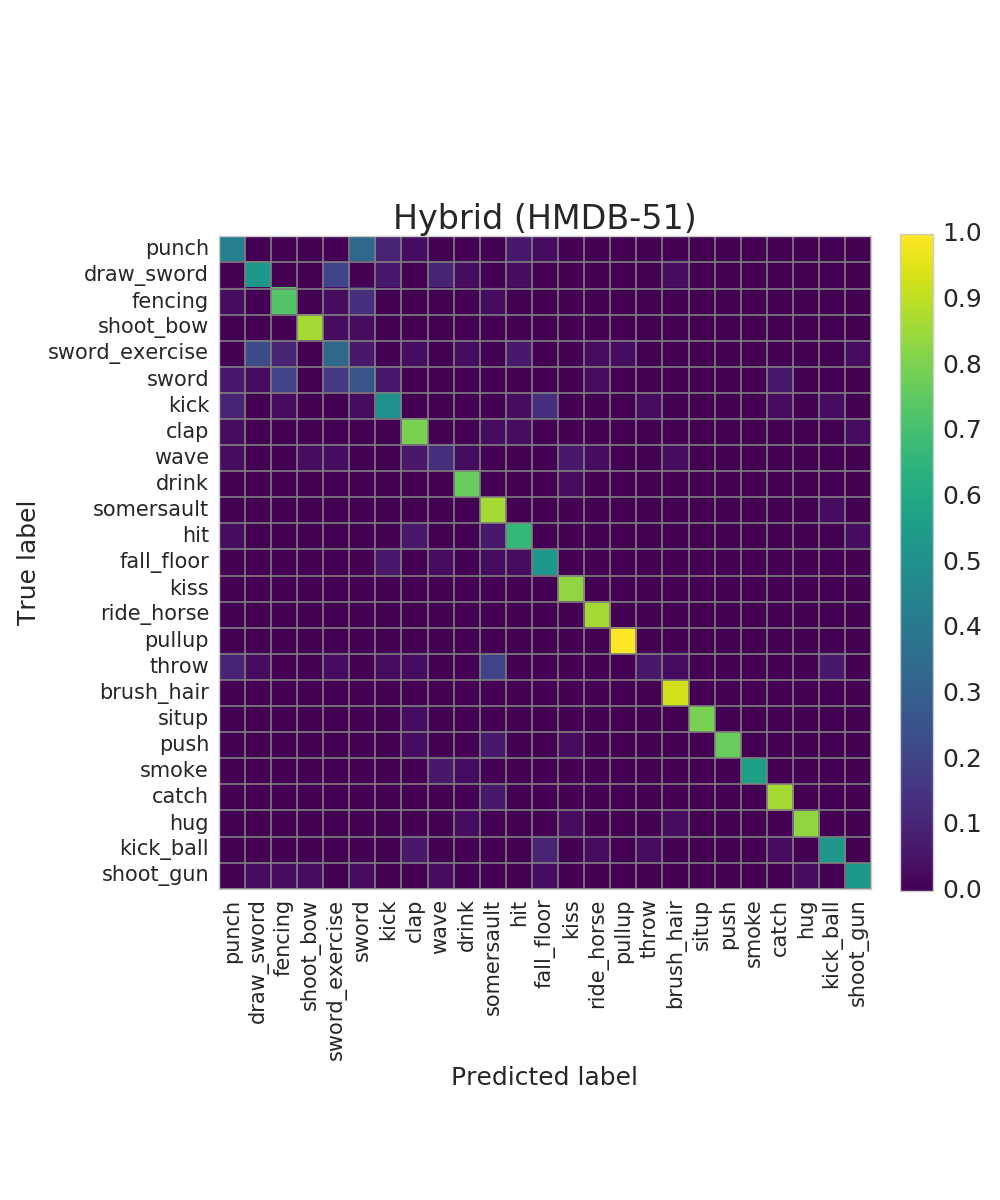}
    \end{subfigure}
    \vspace{-1.0cm}
    \caption{Top-25 most confused classes for HMDB-51.}
    \label{fig:cm_hdb}
    \vspace{-1.2cm}
\end{figure}

HMDB-51 is the second largest dataset we investigate. Using split 1, we
generate the confusion matrices using the same method as before and show the
evolution of the top 25 classes in Figure~\ref{fig:cm_hdb}.

Most of the confusion in this dataset originates from the classes
\textit{punch}, \textit{draw\_sword}, \textit{fencing}, \textit{shoot\_bow},
\textit{sword\_exercise} and \textit{sword}. While our Hybrid model
significantly improves the mean accuracy for this dataset ($+2.6\%$, \cf Table
8 in the main text), the improvements are well distributed across all classes
and are less visible when inspecting individual class pairs. Some
improvements are between the classes \textit{sword\_exercise} and
\textit{shoot\_bow}, \textit{kick\_ball} and \textit{catch}, \textit{wave} and
\textit{sword\_exercise}.

\newpage

\subsection{Hollywood2}

\begin{figure}[h!]
    %\centering
    \vspace{-0.7cm}
    \hspace{-0.7cm}
    \includegraphics[width=1.15\linewidth]{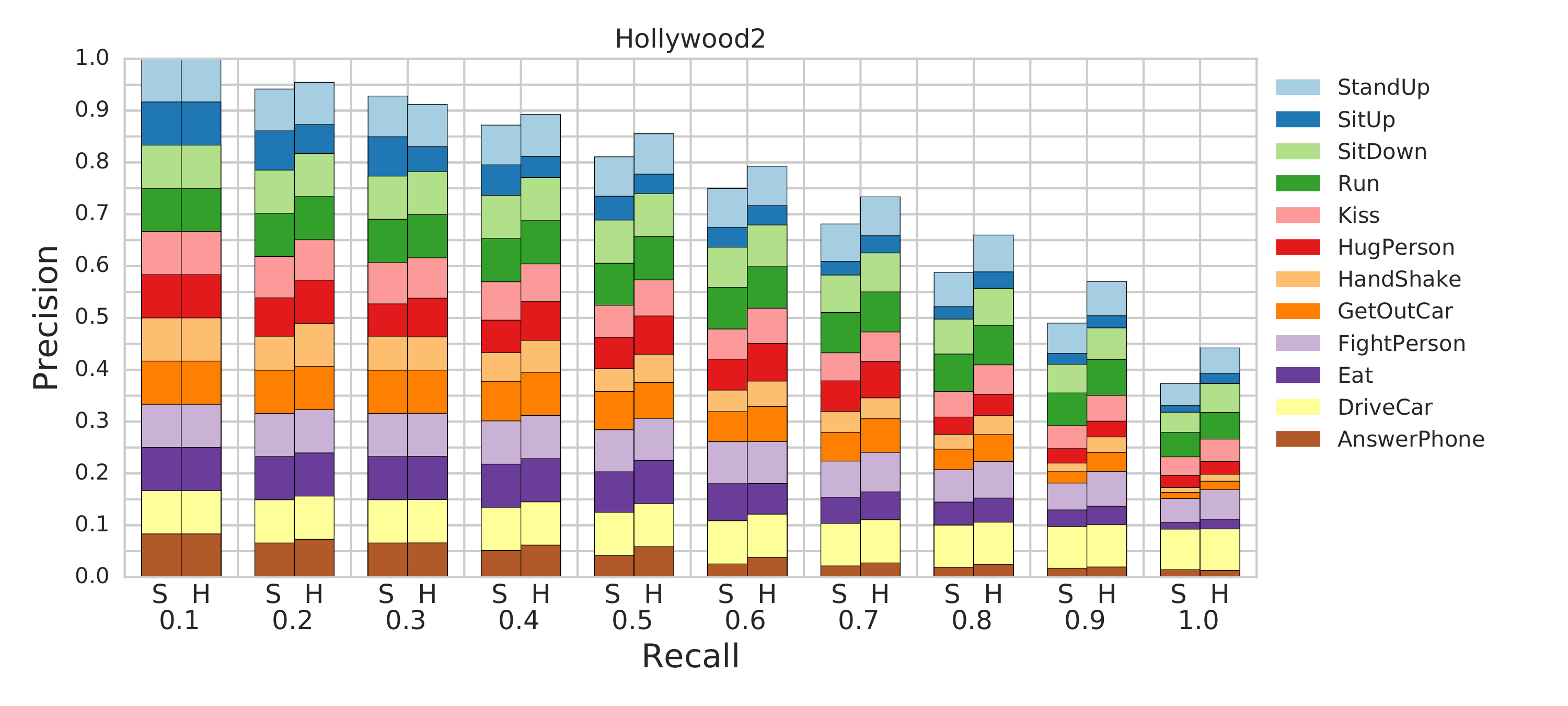}
    \vspace{-1.0cm}
	\caption{Quantized precision and recall segmented per class for the
	Hollywood2 dataset, comparing class-specific precision between our baseline
	model (S) and the best hybrid model (H) at different recall rates.}
    \label{fig:pr_hw2}
    \vspace{-1.0cm}
\end{figure}

Hollywood2 is the third largest dataset we investigate. As it is multi-label,
we present a cumulative quantized precision-recall bar chart segmented
per-class in Figure~\ref{fig:pr_hw2}.
This visualization can be understood as a quantized version of the
precision-recall curve, but allows us to identify the separate influence of
each class in the performance result. 

We can see from the figure how the stacked bars corresponding to the Hybrid 
model (H) associated with higher recall rates present higher precision than the 
baseline (S). The difference in performance comes mostly from 
the \textit{GetOutCar} and \textit{HandShake} classes, whose precision improves at 
higher recall rates for the Hybrid model.

\newpage

\subsection{Olympics}

\begin{figure}[h!]
    \vspace{-2.0cm}
    \hspace{-0.5cm}
    \begin{subfigure}{0.555\textwidth}
        \centering
        \includegraphics[width=1.0\linewidth,
        trim={0 0 2.7cm 0},clip]{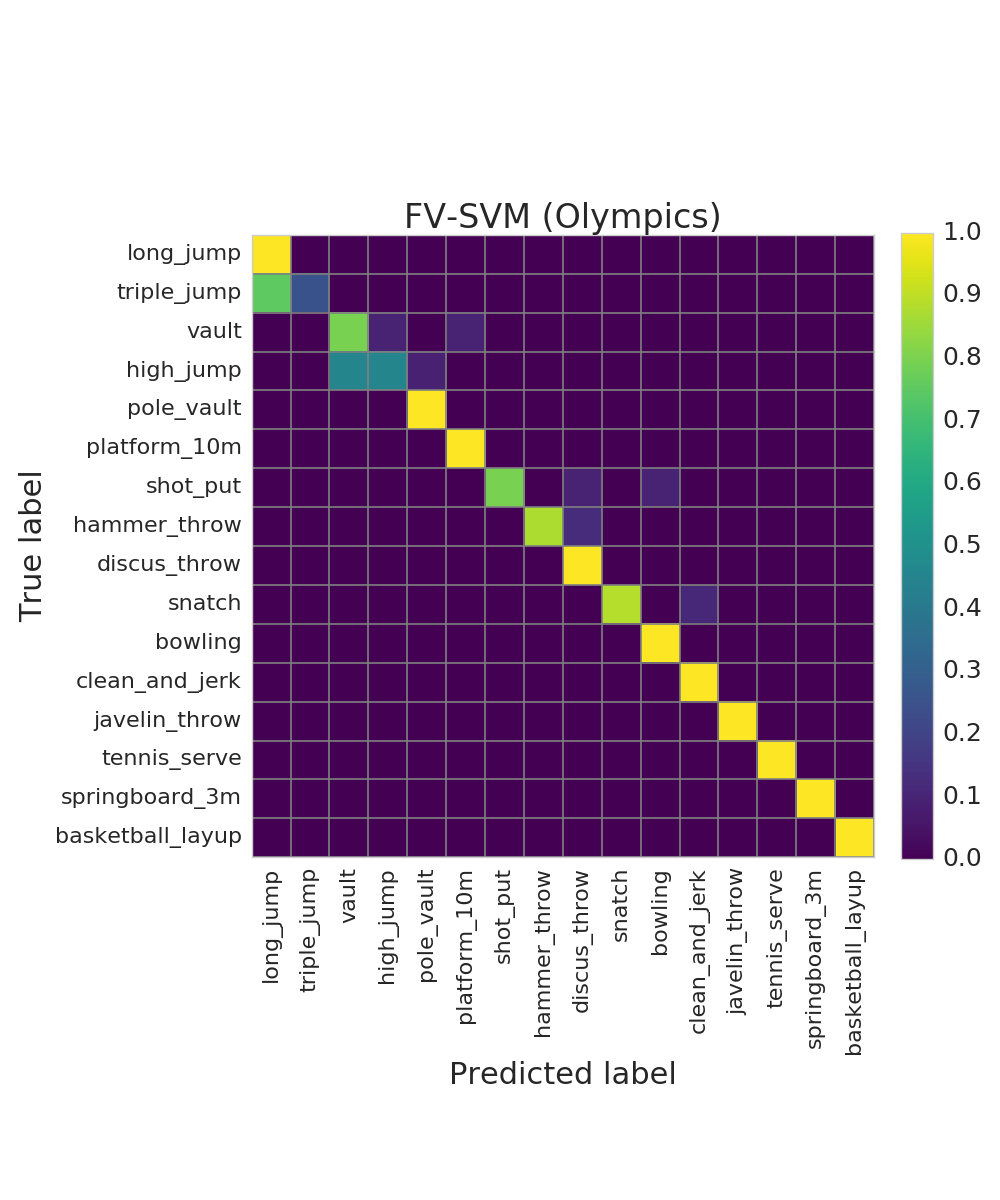}
    \end{subfigure}
    \begin{subfigure}{0.465\textwidth}
        \centering
        \includegraphics[width=1.0\linewidth,
        trim={6.3cm 0 0 0},clip]{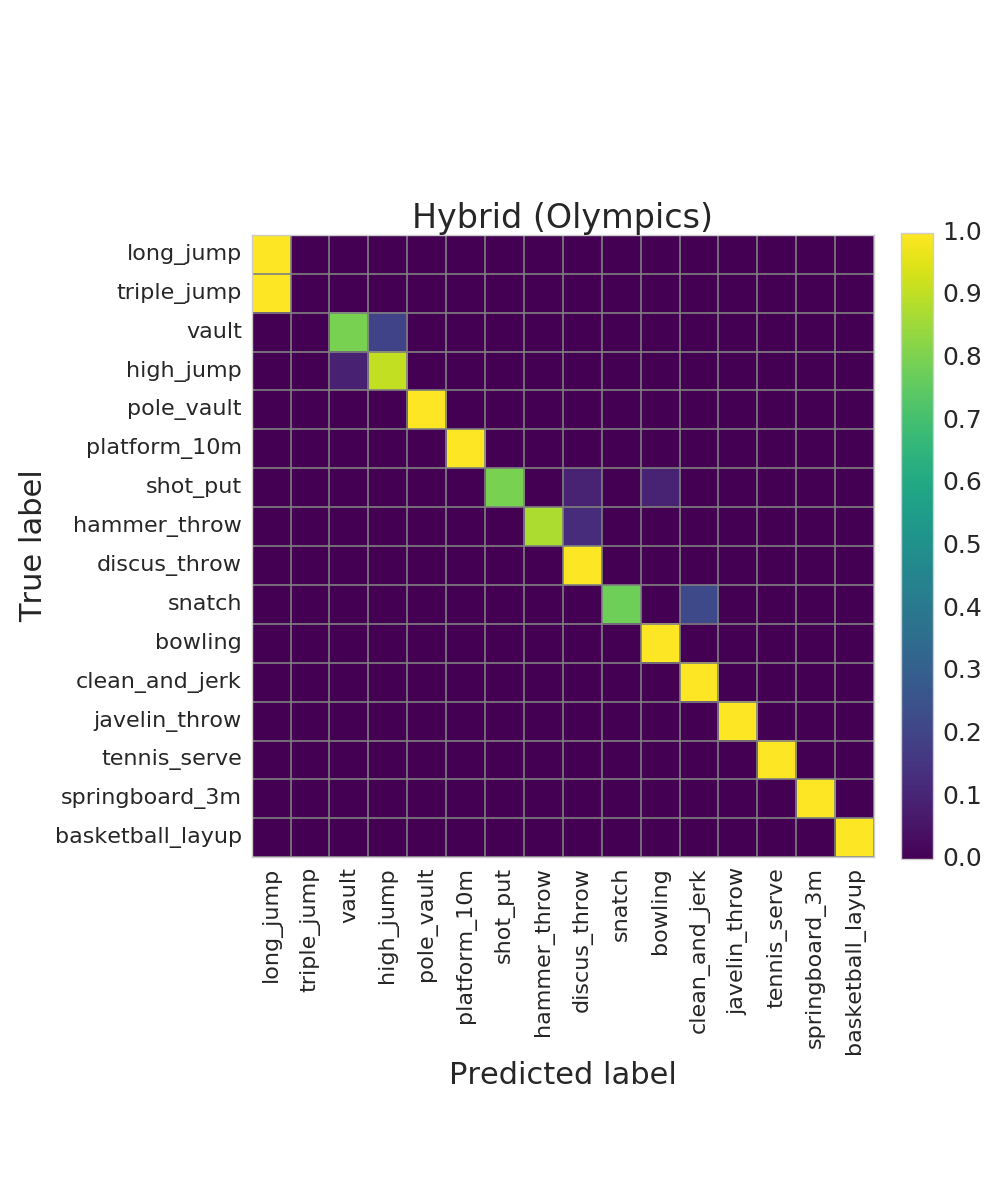}
    \end{subfigure}
    \vspace{-1.0cm}
    \caption{Confusion matrices for Olympics.}
    \label{fig:cm_oly}
    \vspace{-1.4cm}
\end{figure}

\begin{figure}[h!]
    \hspace{-0.7cm}
    %\centering
    \includegraphics[width=1.12\linewidth]{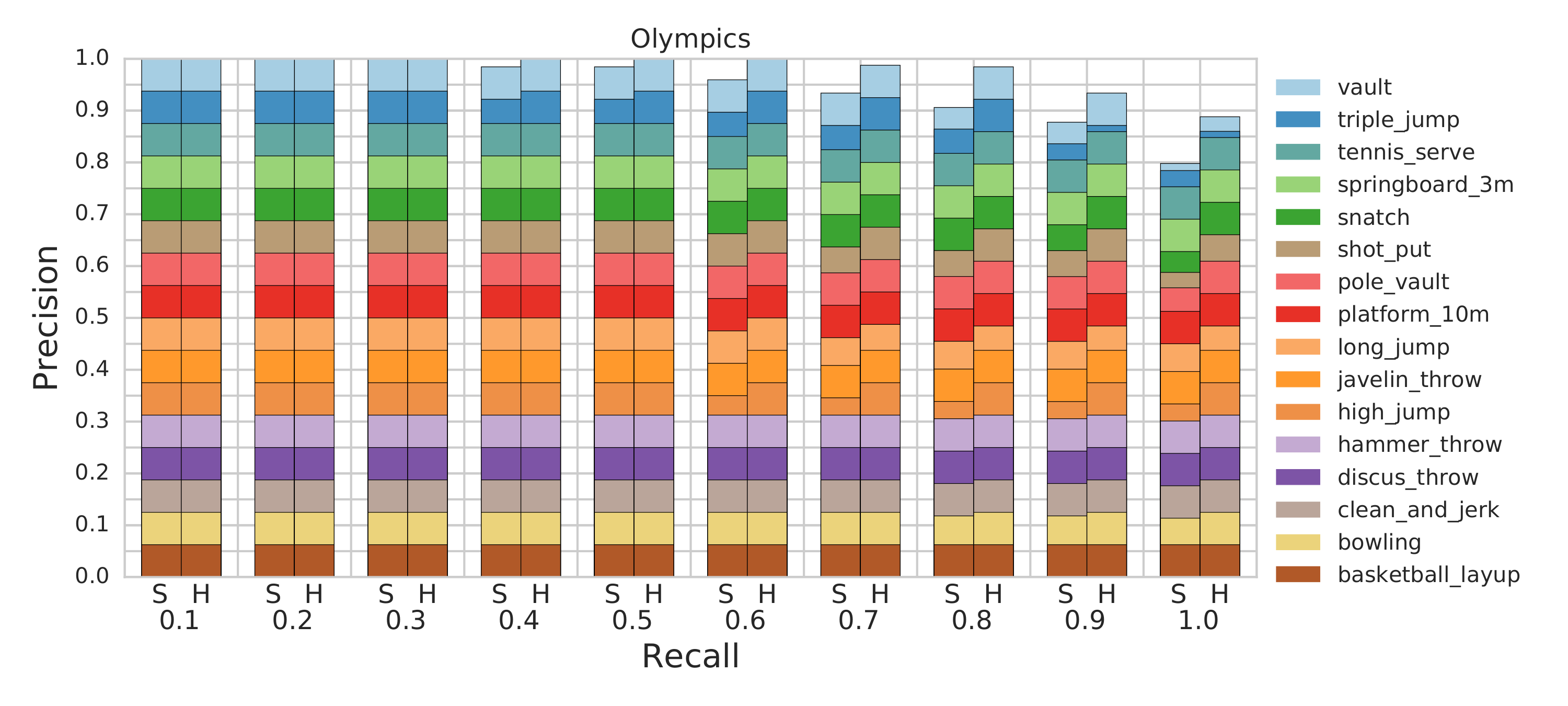}
    \vspace{-0.9cm}
    \caption{Quantized precision-recall segmented per class for Olympics.}
    \label{fig:pr_oly}
    \vspace{-1.0cm}
\end{figure}

Olympics is multi-class and evaluated with mAP, so we present both confusion
matrices (Figure~\ref{fig:cm_oly}) and per-class quantized precision-recall bar
charts (Figure~\ref{fig:pr_oly}).
Most of the confusion originates from the \textit{long\_jump},
\textit{triple\_jump}, \textit{vault}, and \textit{high\_jump} classes. Our
hybrid classifier is able to solve the strong confusion between
\textit{high\_jump}, \textit{vault}, and \textit{pole\_vault}, as well as
\textit{vault} and \textit{platform\_10m}.
However, the Hybrid model has trouble distinguishing between
\textit{long\_jump} and \textit{triple\_jump}, the smallest class in the
dataset (17 videos for training, only 4 for testing). It has therefore low
impact on mean accuracy:  the overall performance in this dataset is indeed
significantly improved ($+3.9\%$ \wrt FV-SVM and $+5.3\%$ \wrt the state of the
art, \cf Table 8 in the main text).

\newpage
Figure~\ref{fig:pr_oly} shows that both methods have excellent precision at low
recall, with the performance between the two methods starting to differ after a
recall of $40\%$. For higher recall rates, we can see how most of the
differences stem indeed from the \textit{triple\_jump} class.
The decrease in area for segments of this class are compensated by steady
improvements in all other classes, as can be seen by comparing the
class-specific segments between the baseline (S) and Hybrid (H) results for the
0.9 and 1.0 recall rates. This suggest that the majority of the improvements
remaining are for this single class.

\vspace{-2mm}
\subsection{High-Five}

\begin{figure}[h!]
	\centering
    \vspace{-1.4cm}
    \begin{subfigure}{0.39\textwidth}
        \centering
        \includegraphics[width=1.0\linewidth,
        trim={0 0 3cm 0},clip]{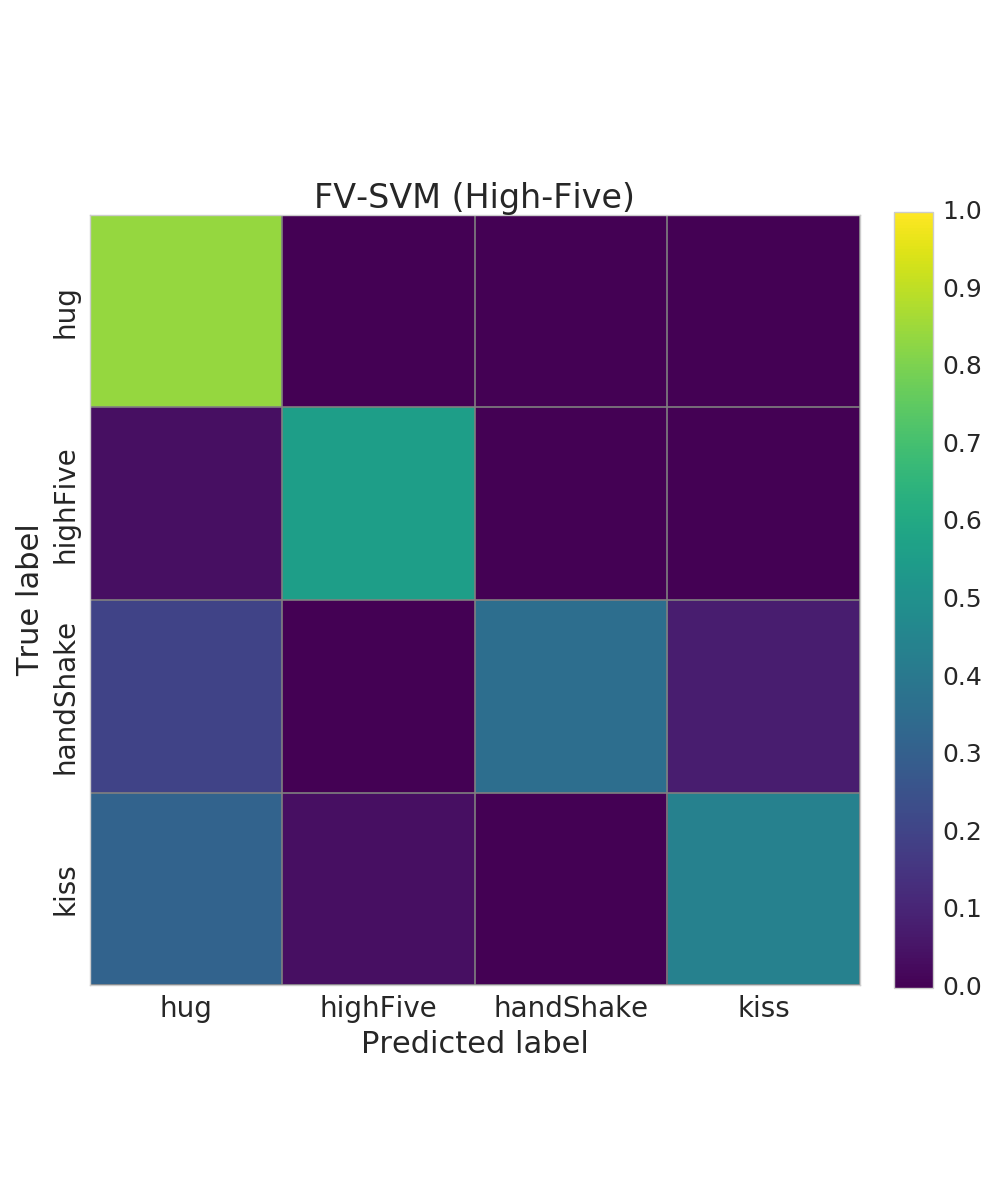}
    \end{subfigure}
    \begin{subfigure}{0.42\textwidth}
        \centering
        \includegraphics[width=1.0\linewidth,
        trim={1cm 0 0 0},clip]{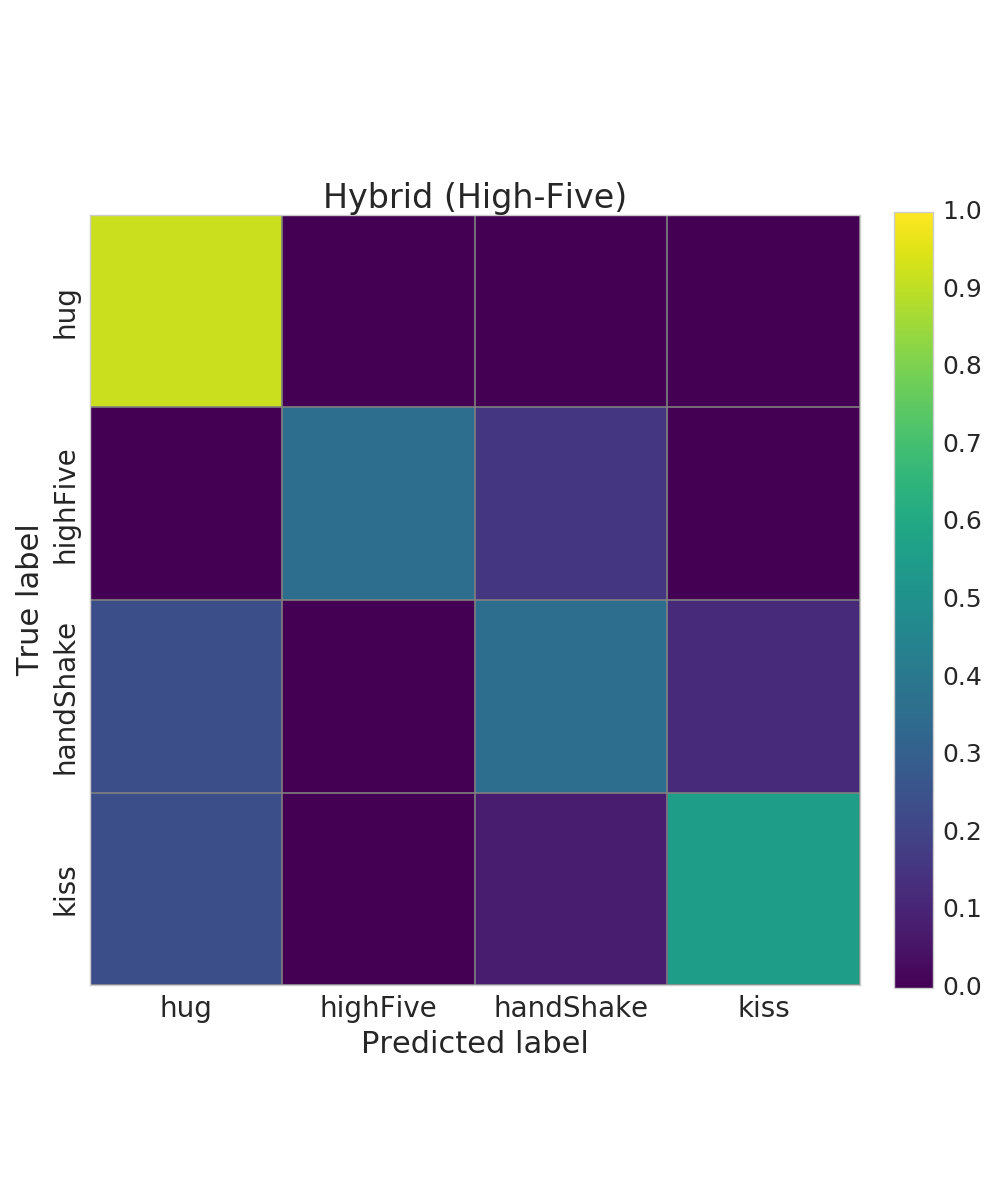}
    \end{subfigure}
    \vspace{-1.0cm}
    \caption{Confusion matrices for the first cross-validation split of 
        High-Five.}
    \label{fig:cm_hi5}
    \vspace{-1.2cm}
\end{figure}

\begin{figure}[h!]
    \vspace{-0.3cm}
    \centering
	%\hspace*{-2mm}
    \includegraphics[width=1.05\linewidth]{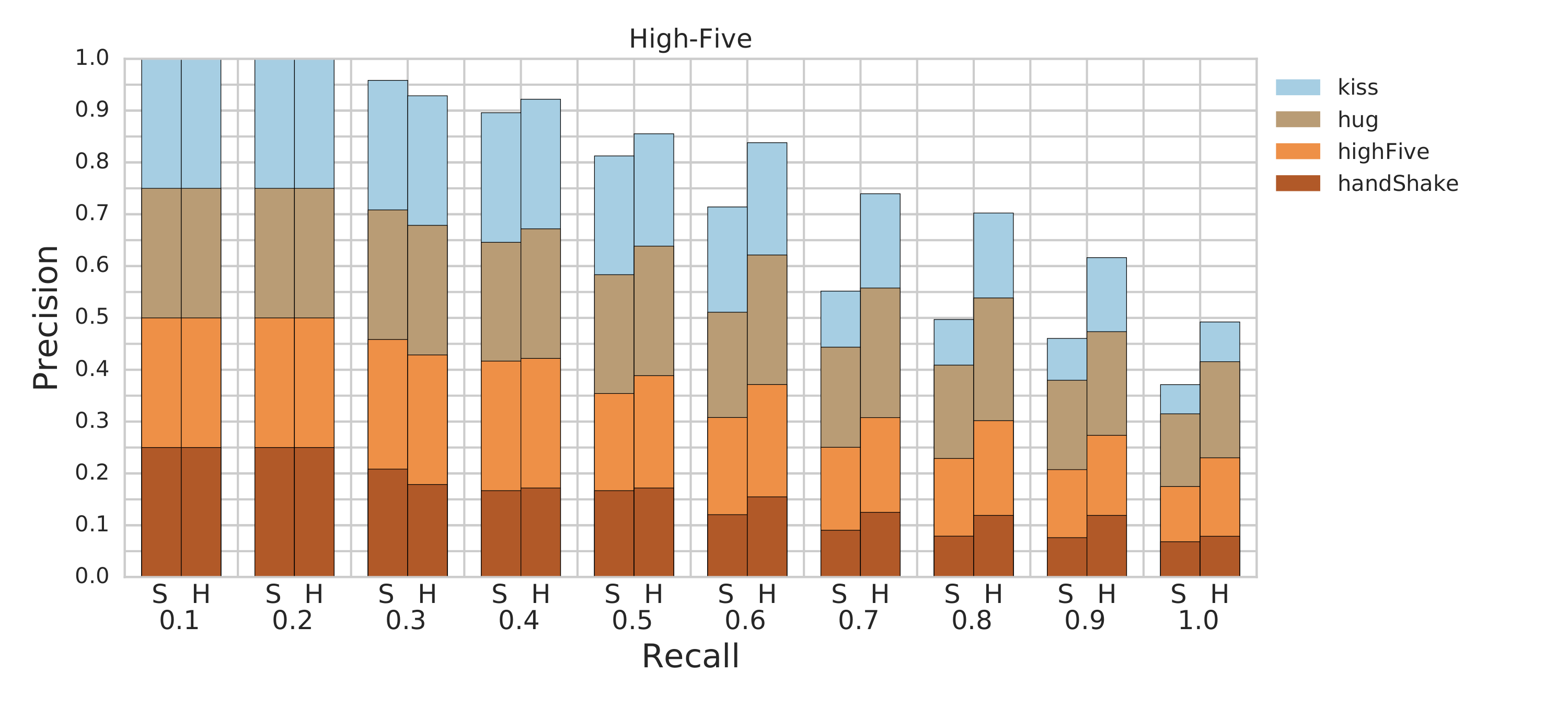}
    \vspace{-0.9cm}
    \caption{Quantized precision-recall segmented per class for split 1 of High-Five.}
    \vspace{-1.5cm}
    \label{fig:pr_hi5}
\end{figure}

High-Five is the smallest dataset (and among the most fine-grained) we
experiment on. As it is multi-class and evaluated with mAP, we present
confusion matrices (Figure~\ref{fig:cm_hi5}) and per-class quantized
precision-recall curves (Figure~\ref{fig:pr_hi5}) on its first cross-validation
split.
Most of the confusion is between \textit{kiss} \vs \textit{hug}, and
\textit{handShake} \vs \textit{hug}. The Hybrid model improves the
disambiguation between \textit{kiss} and \textit{hug}, and \textit{kiss} and
\textit{highFive}, but introduces some confusion between \textit{highFive} and
\textit{handShake}, and \textit{kiss} and \textit{handShake}.
Figure~\ref{fig:pr_hi5} shows that our Hybrid model is more precise at high
recall rates, especially for \textit{kiss} and \textit{handShake}. Moderate
improvements can be observed for all classes.

\newpage

\subsection{Conclusion}

Our detailed per-dataset analysis suggests that the improvements brought by our
hybrid models are generally distributed across the classes of most datasets,
without a single class being responsible for the main performance jump. We also
show that the Hybrid model can differentiate between some fine-grained action
groups, confirming that the unsupervised video-level FV representation contains
fine-grained information about the original video, an information that may
be more successfully exploited by our more complex (deeper) hybrid models.

The experimental evidence presented here therefore strengthens the overall good
results reported in Table 7 of the main text, where we show that our method
improves over the strong FV-SVM baseline between $+1.9\%$ for the large UCF-101
dataset and $+5.7\%$ on the small fine-grained High-Five dataset.

%\bibliographystyle{splncs}
%\bibliography{new}

\end{document}